\documentclass[runningheads]{llncs}

\usepackage[review,year=2024,ID=8806]{eccv}

\usepackage{eccvabbrv}

\usepackage{graphicx}
\usepackage{booktabs}

\usepackage[accsupp]{axessibility}  

\usepackage[pagebackref,breaklinks,colorlinks,citecolor=eccvblue]{hyperref}

\usepackage{orcidlink}

\begin{document}

\title{Towards Better Text-to-Image Generation Alignment via Attention Modulation} 

\titlerunning{Better T2I Alignment via Attention Modulation}

\author{
Yihang Wu*\inst{1} \and
Xiao Cao*\inst{1} \and
Kaixin Li\inst{1} \and
Zitan Chen \inst{2} \and
Haonan Wang \inst{1} \and
Lei Meng \inst{2} \and
Zhiyong Huang \inst{1}
}

\authorrunning{F.~Author et al.}

\institute{National University of
Singapore, Singapore, Singapore \and Shandong University, Shandong, China}

\maketitle

\begin{abstract}
In text-to-image generation tasks, the advancements of diffusion models have facilitated the fidelity of generated results. 
However, these models encounter challenges when processing text prompts containing multiple entities and attributes. The uneven distribution of attention results in the issues of entity leakage and attribute misalignment. Training from scratch to address this issue requires numerous labeled data and is resource-consuming. 
Motivated by this, we propose an attribution-focusing mechanism, a training-free phase-wise mechanism by modulation of attention for diffusion model. One of our core ideas is to guide the model to concentrate on the corresponding syntactic components of the prompt at distinct timesteps. 
To achieve this, we incorporate a temperature control mechanism within the early phases of the self-attention modules to mitigate entity leakage issues. An object-focused masking scheme and a phase-wise dynamic weight control mechanism are integrated into the cross-attention modules, enabling the model to discern the affiliation of semantic information between entities more effectively. 
The experimental results in various alignment scenarios demonstrate that our model attain better image-text alignment with minimal additional computational cost. The code will be released upon the acceptance.

  \keywords{Image Generation \and Diffusion Model \and Attention Control}
\end{abstract}


\section{Introduction}
\label{sec:intro}

In recent years, generative AI techniques made unprecedented advancements across numerous domains. The emergence of large-scale pre-trained models has sparked novel applications in various downstream tasks. This is particularly evident in the area of text-to-image generation, where models such as  Stable Diffusion~\cite{rombach2022high}, DALL-E 2~\cite{ramesh2022hierarchical}, Imagen~\cite{saharia2022photorealistic} have prominently demonstrated their capabilities. In addition to these specialized text-to-image models, the recently unveiled large-scale QA model GPT-4 Vision~\cite{openai2023gpt4} and the video generation model Sora~\cite{liu2024sora} have also shown remarkable text-to-image generation abilities. Nonetheless, challenges arise with complex prompts that include multiple entities and intricate attributes. 
The quality of the generated images declines, resulting in issues such as \textbf{entity leakage}\cite{wang2023compositional}, where objects disappear or their outline break due to entities' positional fusion, and \textbf{attribute misalignment}~\cite{wang2023decompose, wen2023improving, feng2023ernie}, where object's attributes are incorrectly bound
.
Consequently, users are often required to repeatedly refine their prompts and emphasize the adjectives, or utilize additional auxiliary images to obtain the desired visual results.


Currently, most state-of-the-art models employ pre-trained language models to encode the prompt, and subsequently integrate the text embeddings into attention modules, where Gaussian noise is diffused into the final generated image. However, recent studies~\cite{hong2023improving,chefer2023attend,kim2023dense,he2023localized} have identified that many issues with sub-optimal generated images are intricately associated with deficiencies in the attention mechanism. In attention modules, insufficient attention given to the queried object may result in object omission in the final result. Conversely, a conflated attention distribution across multiple objects on the attention maps can result in entity leakage and fusion errors. Regarding attribute alignment, issues often arise when the attention response for an attribute inadvertently spreads into the region of another object, disrupting the intended alignment. 

Previous methods~\cite{xie2023boxdiff, li2023gligen, yang2023reco, chai2023layoutdm, cheng2023layoutdiffuse, hui2023unifying, xie2023boxdiff, inoue2023layoutdm,avrahami2023spatext} relied on users providing additional explicit layout images to help delineate the position of attributes and entities. This strategy is a viable alternative approach; however, it falls short in tackling unique challenges inherent to \textbf{text-to-image} synthesis tasks. Some studies ~\cite{chefer2023attend, wang2023compositional,feng2022training} have concentrated on enhancing attention control. However, most of these strategies are designed for specific cases such as compositional text-to-image tasks without undergoing extensive testing in more general generative contexts. Additionally, some come with trade-offs, such as significant increases in computational costs. 

\begin{figure*}[!t] 
  \centering
  \includegraphics[width=0.9\textwidth]{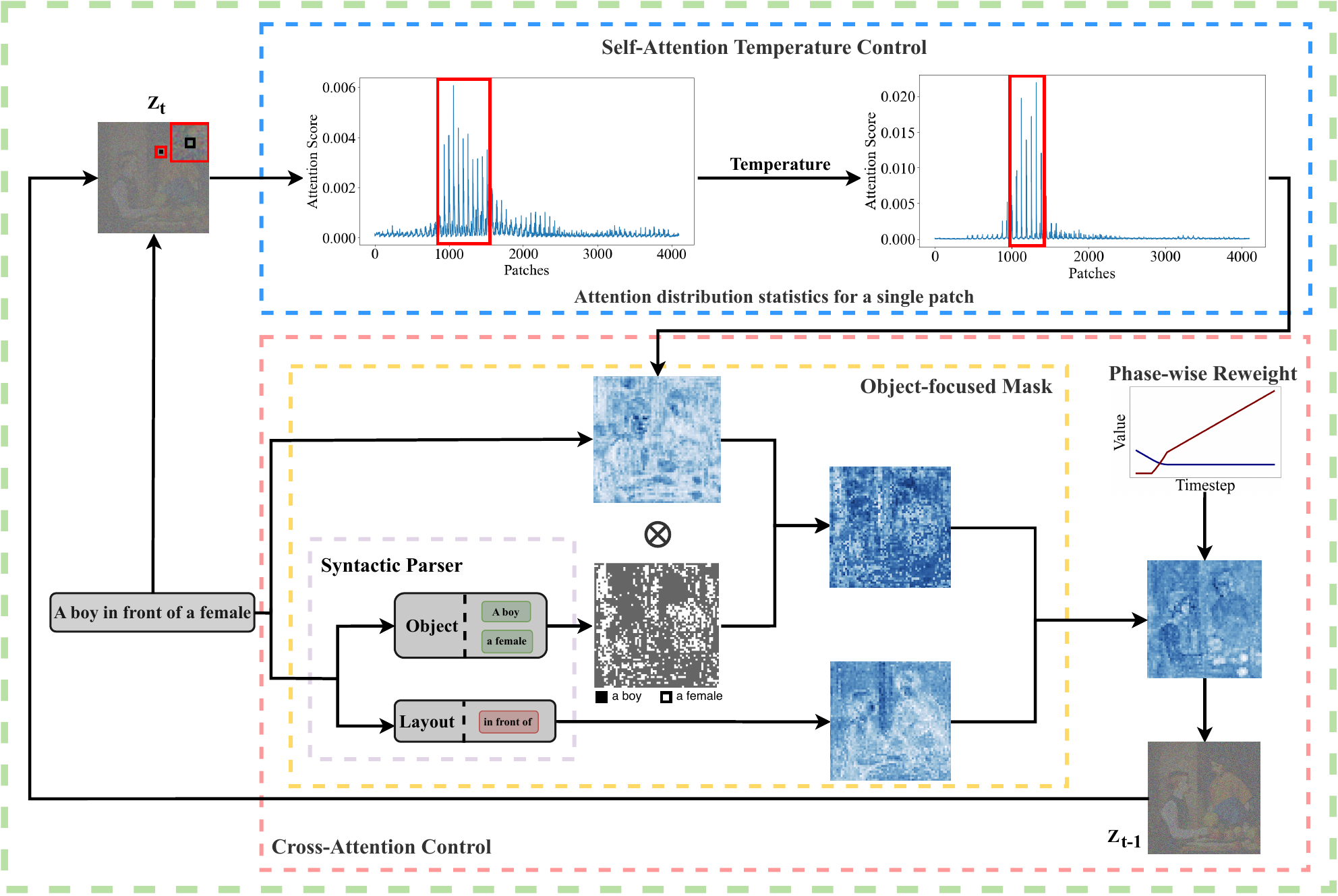} 
  \caption{\textbf{The overall pipeline of our methods.} In the self-attention module, we employ a temperature control strategy to better construct the outlines of the entities. In the cross-attention layers, we integrate an object-focused masking mechanism and a dynamic reweighting mechanism to emphasize different components of the prompt at various stages.}
  \label{pipeline}
\end{figure*}

In this work, we propose an efficient training-free phase-wise attention control paradigm to alleviate entity leakage and attribute misalignment.
Our proposed methodology includes three parts: \textbf{1) Self-attention temperature control}, where we modulate the temperature in self-attention to mitigate entity leakage issues, with which improved entity boundaries are observed. \textbf{2) Object-focused cross-attention mask}, where the leaked attention carrying semantic information from tokens of other unrelated objects and attributes are masked for every patch.
By ensuring that each patch focuses exclusively on a single entity group, the occurrence of attribute misalignment is significantly reduced. \textbf{3) Phase-wise dynamic reweighting strategy} is further proposed to improve attribute alignment by varying the emphasis on different semantic components of the prompt at various stages during the generation process. The main contributions of our work can be summarized as:

\begin{itemize}
\item We propose a training-free phase-wise attention control paradigm to improve the issues of entity leakage and attribute misalignment. A self-attention temperature control paradigm is implemented. To the best of our knowledge, We are the first to mitigate the entity leakage issue through self-attention temperature control.
\item A novel object-focused masking scheme and a dynamic reweighting mechanism within cross-attention layers are proposed, enabling the model to prioritize various semantic components of the prompt during different generating stages.
\item We conduct metric-evaluated and semi-human-evaluated experiments and ablation experiments. The experimental results show that our approach achieves state-of-the-art results on both qualitative metrics and quantitative results.
\end{itemize}

\section{Related Work}
\label{related}
\textbf{Diffusion-based models}
The diffusion model~\cite{ho2020denoising} has achieved significant success in various content generation fields, the core concept of which involves iteratively reconstructing images from noise in the latent space through the diffusion processes guided by the input prompt. Existing diffusion-based image generation models~\cite{ ramesh2022hierarchical, saharia2022photorealistic} are trained on massive datasets and have demonstrated considerably improved performance compared to previous methods. However, when dealing with complex input prompts, the fidelity of these generation models often cannot be guaranteed~\cite{zhang2023adding, ruiz2023dreambooth,saharia2022photorealistic}. Recently, the diffusion-transformer-based text-to-video model Sora~\cite{liu2024sora} has been unveiled, showcasing strong capabilities in generating high-fidelity, long-duration, high-resolution videos. Furthermore, it has also been announced to possess robust text-to-image generation capabilities. However, due to the non-disclosure of its model architecture and weights, the detailed performance of its text-to-image generation quantitative performance results remain unassessable, and we also cannot use the software for comparative study. 

\textbf{Diffusion-based image synthesis} 
Given the substantial computational resources required for training diffusion-based models, current research primarily focuses on enhancing existing large-scale models. This enhancement is primarily explored through two main directions: layout-to-image(L2I) generation~\cite{chai2023layoutdm, cheng2023layoutdiffuse, hui2023unifying, xie2023boxdiff, inoue2023layoutdm,avrahami2023spatext} and text-guided image(T2I) generation~\cite{chefer2023attend, wang2023compositional, feng2022training, phung2023grounded, hertz2022prompt, kumari2023multi,balaji2022ediffi,zhang2023adding}. In the L2I scenario, the layout is explicitly provided as an auxiliary input condition in the form of bounding boxes, segmentation maps, or similar representations. This layout information actively participates in the diffusion process, guiding image generation. 
In the T2I scenario, textual descriptions serve as the model's sole input. The model is equipped with the capability to internally predict layouts or generate layouts more effectively based on the provided text. 
Although providing additional layout constraints can significantly enhance the realism of generated images, this requires users to have a more detailed conception of the expected image, thereby increasing their workload. Moreover, predefining the overall layout of an image can also limit the model's creativity to some extent. 

\textbf{Attention control-based text-to-image synthesize} Recently, some attention control strategies~\cite{hong2023improving,chefer2023attend,kim2023dense,he2023localized,wang2023compositional,feng2022training,phung2023grounded} have been proposed to enhance T2I generation. Attend-and-excite~\cite{chefer2023attend} calculates loss on the smoothed attention map at each timestep and updates attention values according to gradients to emphasize the tokens that are most easily overlooked. Wang et al. trained BoxNet~\cite{wang2023compositional} to predict the positions of entities obtained from syntactic analysis of the prompt and then generated location masks to create layouts. Feng et al. proposed Structured Diffusion~\cite{feng2022training}, where the attention of different entities in the prompt is computed separately and then superimposed onto the corresponding token positions in the original attention map, to emphasize these entities for better compositional T2I generation. Phung et al.~\cite{phung2023grounded} introduced re-focusing losses for both self-attention and cross-attention mechanisms to improve the alignment between text and layout while reducing the model's attention on less important areas. Among these methods, some ~\cite{wang2023compositional,feng2022training,chefer2023attend} focus solely on generating images for compositional prompts without testing in general text generation scenarios. 

\section{Proposed Method}
\label{method}
In this work, we propose an efficient training-free attention control paradigm by modulating the self-attention and cross-attention modules. Our scheme's overall pipeline is illustrated in Figure \ref{pipeline}. In the self-attention layer, the attention distribution is rescaled wih a temperature operation in the early stage of the diffusion process to guide the model to address the entity leakage problem, as depicted in Section \ref{sec31}. Subsequently, we delve into the object-focused masking mechanism and dynamic reweighting mechanism in cross-attention layers designed to resolve attribute misalignment, which will be discussed in Section \ref{sec32}. These two mechanisms allow each patch to focus primarily on one instance and emphasize different semantic components of the prompt during the corresponding phase.
Our method achieves better performance with nearly no additional time cost. The average time cost for generating one image for ours and SDXL is 28.94s and 28.50s (Our/SDXL=101.54\%) on one NVIDIA TITAN RTX in our experiment study.

\begin{figure*}[h!]
    \centering
    \begin{subfigure}[b]{0.65\textwidth} 
        \includegraphics[width=\textwidth]{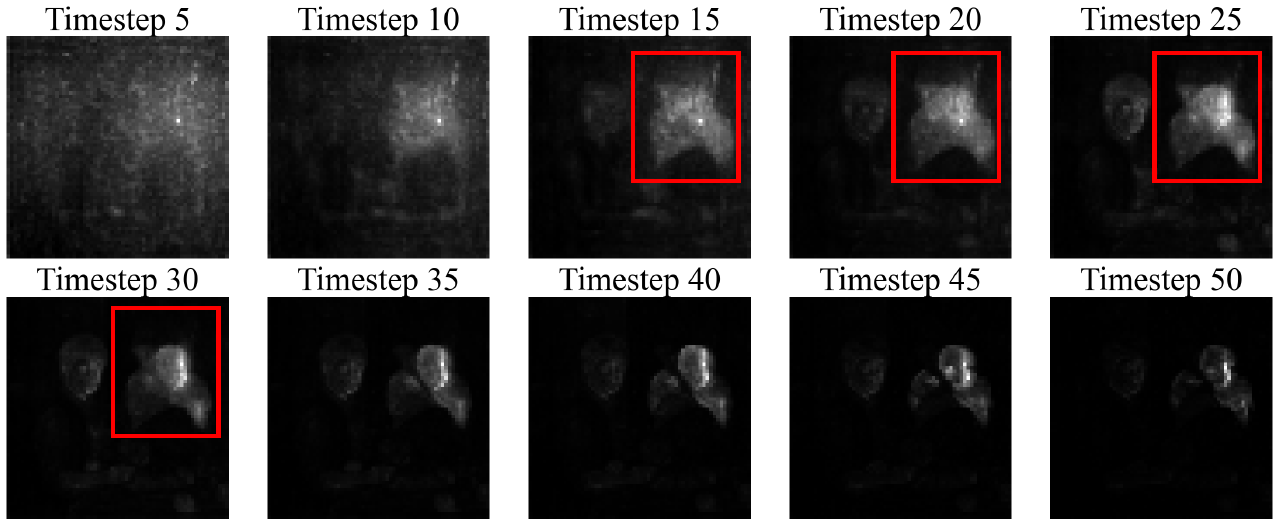}
        \caption{The response of the patch to other patches during diffusion process without temperature control.}
        \label{ori_self_vis}
    \end{subfigure}
    \hfill 
    \begin{subfigure}[b]{0.25\textwidth} 
        \includegraphics[width=\textwidth]{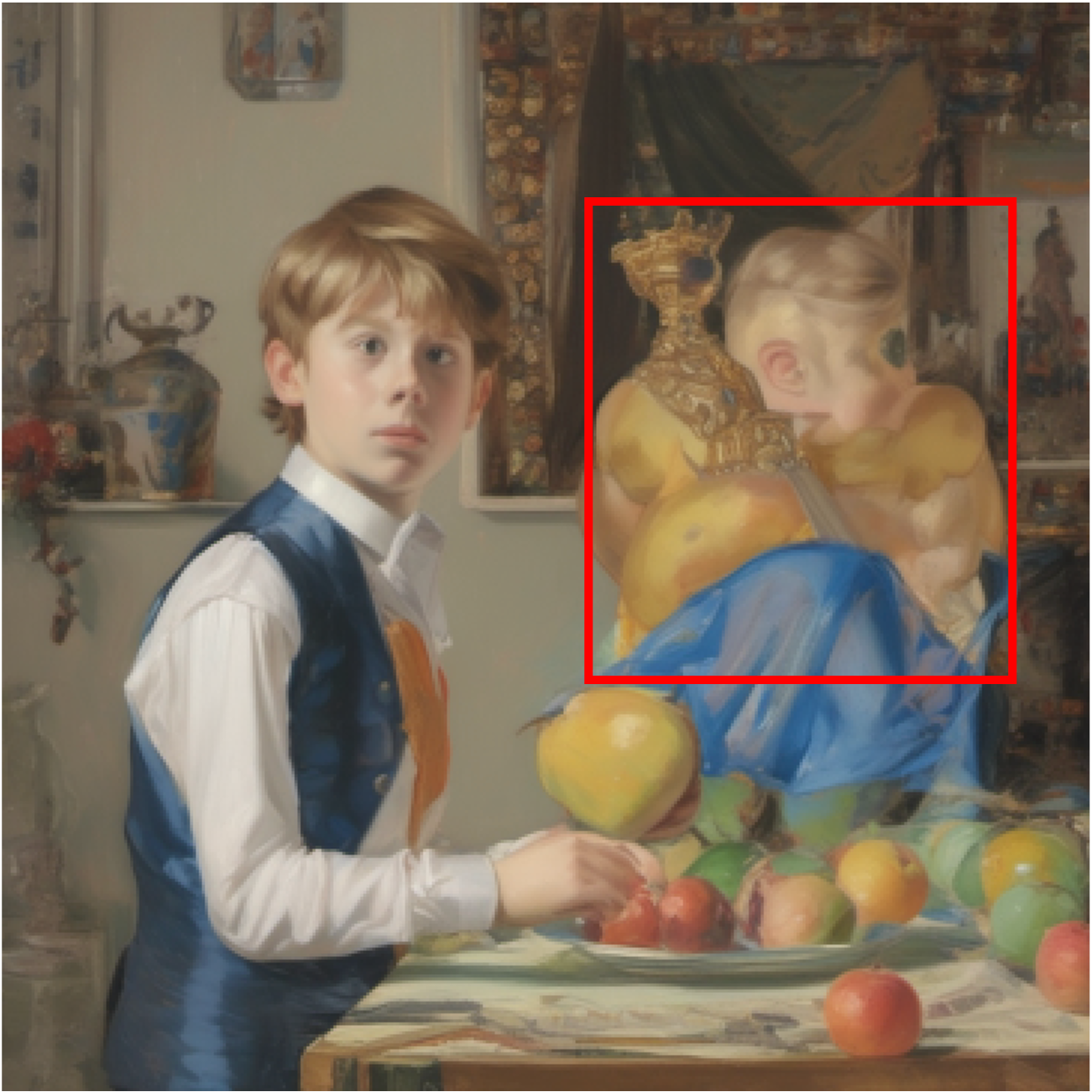}
        \caption{Original generated image}
        \label{ori_with_box}
    \end{subfigure}

    \vspace{0.5cm} 

    \begin{subfigure}[b]{0.65\textwidth}
        \includegraphics[width=\textwidth]{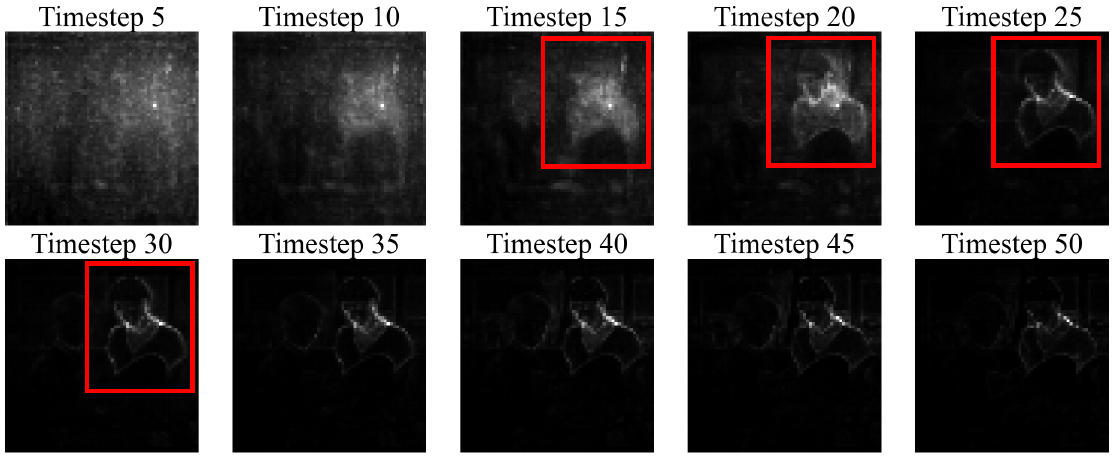}
        \caption{The response of the patch to other patches during diffusion process with temperature control.}
        \label{our_self_vis}
    \end{subfigure}
    \hfill
    \begin{subfigure}[b]{0.25\textwidth}
        \includegraphics[width=\textwidth]{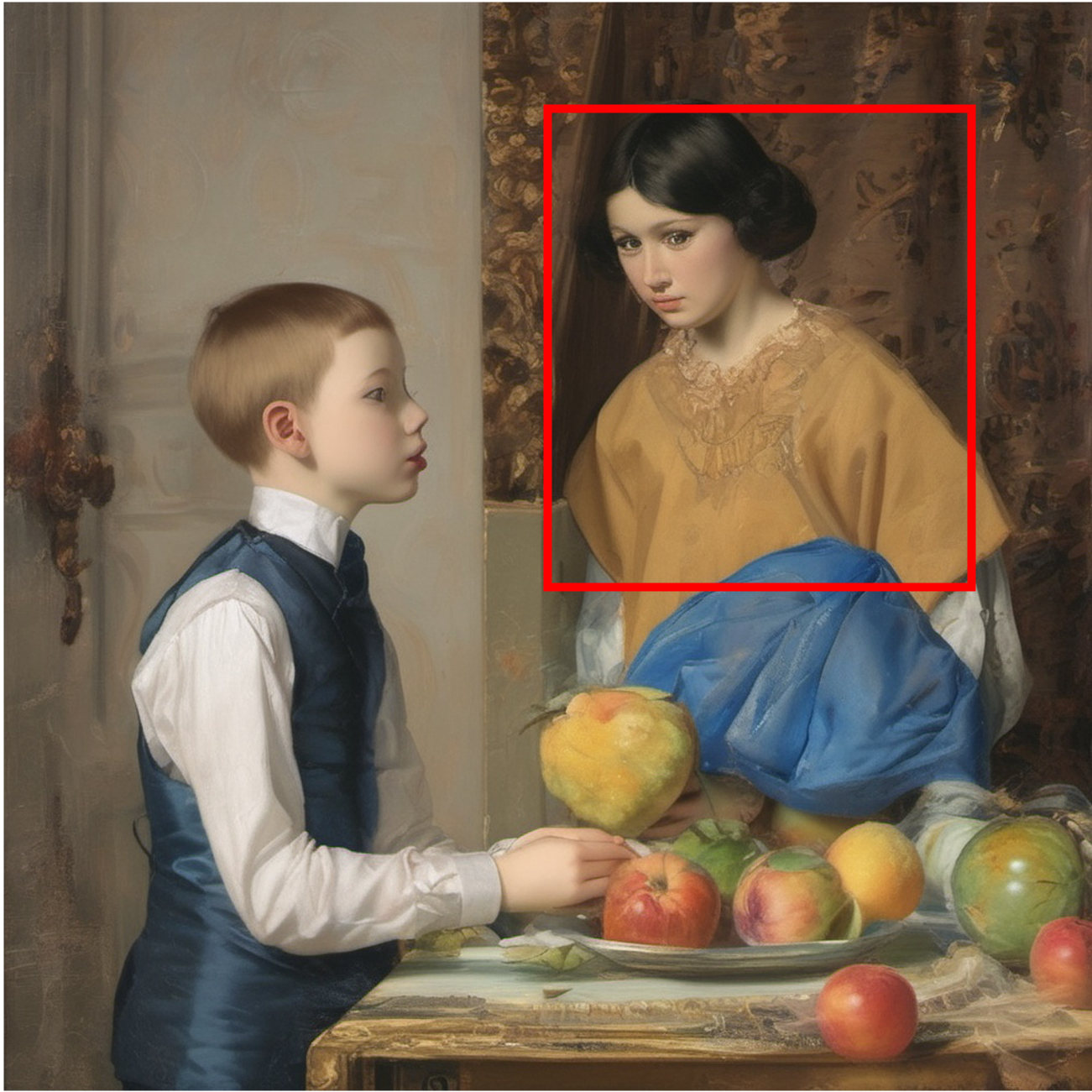}
        \caption{Our generated image}
        \label{our_with_box}
    \end{subfigure}

    \caption{\textbf{A display of the effects of self-attention temperature control on the self-attention map.} Given the prompt "a boy in front of a female",
    we visualized the attention values between a patch within the highlighted red box in figures (b) and (d) and other patches throughout the diffusion process. After applying temperature control, the patch's high response region became more confined, thereby forming more accurate outlines.}
    \label{self_example}
\end{figure*}
\vspace{-20px}
\subsection{Self-Attention Control}
\label{sec31}

Some recent work~\cite{phung2023grounded,chefer2023attend,wang2023compositional} has indicated that improper attention allocated to the image patches in the self-attention layers could lead to erroneous or vague object outlines in the generated image. Such observations align with our experiment results in Figure \ref{ori_self_vis}. For a given patch, if it exhibits high response values with a larger surrounding area, the model may become perplexed and fail to definitively assign the patch to its corresponding entity. 

Hence, we propose an intuitive methodology to modify the distribution of self-attention values for patches through temperature operation as follows:
\begin{equation}
    \text{A}_{\tau}^{s} = \text{softmax}\left(\frac{QK^T}{\tau \sqrt{d_k}}\right)V
\end{equation}
Here, $\tau$ is a hyperparameter used to specify the temperature. Through this temperature operation, high attention values between patches with strong correlations are emphasized, while low attention values between unrelated patches are suppressed. This allows each patch to group with highly correlated patches, resulting in more realistic outlines. 
In Figure \ref{our_self_vis}, after temperature control, the patch only corresponds with patches within a smaller surrounding area, leading to the correct outlines being constructed in the final generated image.
It is essential to note that since self-attention control directly alters the relationships between patches, such control must be precise and moderate; otherwise, it can impact the overall content of the final image. 
Therefore, we apply the temperature operation to the early generation stage of the diffusion model in the self-attention layer. 
\subsection{Cross-Attention Control}
\label{sec32}
\subsubsection{Object-focused masking mechanism}
\begin{figure*}
    \centering
    \begin{subfigure}[b]{1\textwidth}
        \centering
        \includegraphics[width=\textwidth]{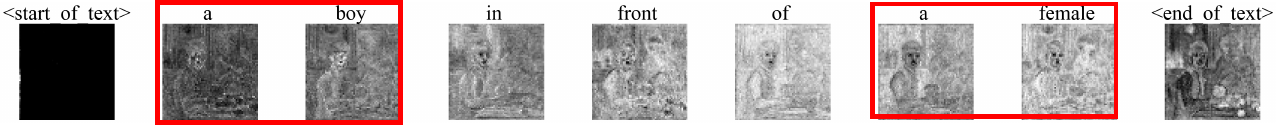}
        \caption{Timestep = 30, the cross-attention map of the original model.}
        \label{cross_ori}
    \end{subfigure}

    \begin{subfigure}[b]{1\textwidth}
        \centering
        \includegraphics[width=\textwidth]{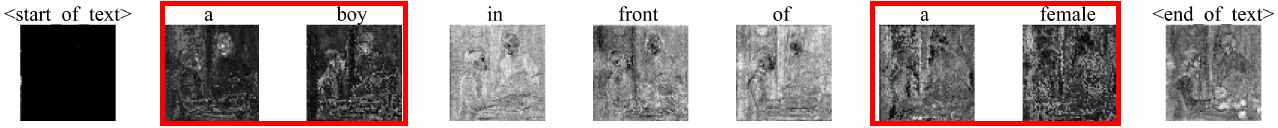}
        \caption{Timestep = 30, the cross-attention map of our model.}
        \label{cross_our}
    \end{subfigure}
    
    \caption{\textbf{A comparison of cross-attention maps of each token original model and our method at timestep 30 with different methods.} Given the prompt of "a boy in front of a female", we can observe that the semantic information of some tokens is spread throughout the entire image in the original model, resulting in poor alignment. With our control method, the token information corresponding to entities becomes more aggregated, ultimately yielding better generative results.}
    \label{cross_comp}
\end{figure*}
\vspace{-10px}
In Figure \ref{cross_comp}, we observed that prompts containing multiple distinct objects often lead to difficulties in accurately defining the boundaries between these objects, which result in an unrealistic generation with vanishing or merging of entities.
 In Figure \ref{cross_ori}, the semantic information of ``female" is spread throughout the entire attention map. This implies that the patch cannot effectively identify the outlines of the female entity, leading to the model's inability to construct the correct entity. Simultaneously, this redundant information, diffused throughout the image, can also affect the construction of other entities, easily leading to issues such as entity leakage. In this case, for a given patch, there may be multiple instances with high attention values on that patch. As the generation process proceeds, the patch needs to represent semantic information for multiple tokens, leading to content mixing. Therefore, we have designed an object-focused masking mechanism to release the redundant information that attention patches are required to attend to.

Specifically, for the input prompt $P$, we use syntactic parsing $f_p(\cdot)$ to identify all entities align with attributes in the prompt, such as "a white cat" or "a young boy with a hat," denoted as \( E = \{e_1, e_2, \ldots, e_m \)\}. Correspondingly, we can obtain other semantic components in the prompt \( O = \{o_1, o_2, \ldots,o_n \)\}, including verbs, predicates, and layout-related information in the image. These phrases' positional ranges in the original sentence are denoted as \( R_E = \{r_{e_1}, r_{e_2}, \ldots, r_{e_m}\} \) for \( E \) and \( R_O = \{r_{o_1}, r_{o_2}, \ldots, r_{o_n}\} \) for \( O \).

Then, for the cross-attention map \( A \in \mathbb{R}^{H \times W \times T} \), where $T$ is the max length of the token sequence, we calculate the probability distribution of all entity groups as follows:
\begin{equation}
     S_E(i) = \sum_{t \in r_{e_i}} A_{h, w, t}, \ \forall i \in E, \ h \in [1, H], \ w \in [1, W] 
\end{equation}
where  $H,W$ are the height and width of the attention map.

Next, for each patch \( (h, w) \), we find the entity group with the highest probability:
\begin{equation}
     M_{h,w} = \arg\max_i S_E(i)
\end{equation}

The current object-focused mask $M_{h,w}$ only retains information from a single entity group on each patch. This means that, apart from other entities, the rest information specified in the prompt, such as layout and actions, is also lost. Therefore, we need to preserve this information related to the overall image layout, so the modified object-focused mask can be represented as:

\[ M_{h, w, t} = 
\begin{cases} 
1 & \text{if } t \in r_{M_{h,w}} \text{ or } t \in R_O \\
0 & \text{otherwise} 
\end{cases}
\]

Finally, we calculate new attention distribution based on the mask:
\begin{equation}
    A'_{h, w, t} = A_{h, w, t} \times M_{h, w, t}
\end{equation}

With this masking mechanism, for each patch, we retain semantic information for only the entity group with the highest probability, along with the global information related to the layout. This approach helps reduce occurrences of object dissolution and misalignment of attributes. In Figure \ref{cross_our}, the attention map for each token, after being masked, becomes much clearer, and the phenomenon of semantic information diffusing throughout the entire image is significantly reduced.

\subsubsection{Phase-wise Dynamic Reweighting}
\label{sec:reweighting}

\begin{figure*}
    \centering
    \begin{subfigure}[b]{1\textwidth}
        \centering
        \includegraphics[width=\textwidth]{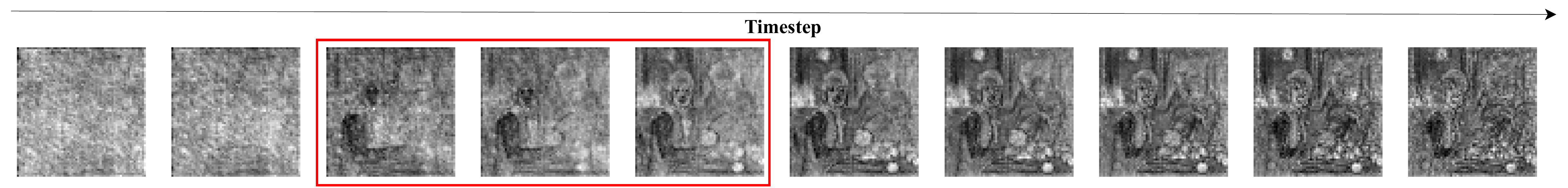}
        \caption{Distribution of attention maps at different timesteps in the original model.}
        \label{cross_ori_weight}
    \end{subfigure}

    \begin{subfigure}[b]{1\textwidth}
        \centering
        \includegraphics[width=\textwidth]{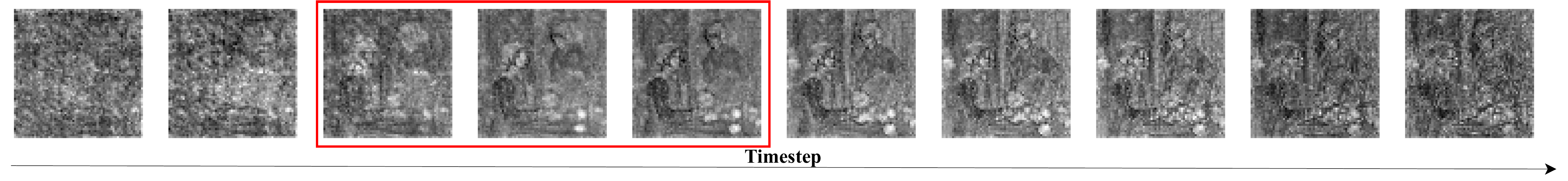}
        \caption{Distribution of attention maps at different timesteps in our model.}
        \label{cross_our_weight}
    \end{subfigure}
    
    \caption{\textbf{Attention maps at different timesteps in original model and our methods.} We visualized the distribution of attention maps at different time steps given the prompt "a boy in front of a female". Compared to the original model, after our control, the attention map can construct the outlines of entities earlier. This is particularly evident in the third and fourth timestep stages, where the dynamic reweighting mechanism allows for earlier and better differentiation between the image background and the entities.}
    \label{cross_comp_weight}
\end{figure*}
\vspace{-10px}
In previous masking mechanism, we divided the prompt into two parts: entities and other semantic information. The former focuses on objects and details, while the latter concentrates on the global layout. This division aligns with the emphasis on different generation stages in diffusion. Some studies~\cite{hertz2022prompt,chefer2023attend} have shown that in the early stages of diffusion, global information like layout is constructed, while in the mid and late stages, details of objects are progressively filled in.

Therefore, on top of the masking, we further designed the phase-wise attention reweighting mechanism. Specifically, we assign different weights to the two masks, controlled by curves with different trends. For the $R_O$ part, which focuses on global information, we use a gradually decreasing curve $ f_{other}(\theta)$ for control, which varies with timestep \( \theta \). For the $R_E$ part, which focuses on instances, we use a gradually increasing curve $f_{other}(\theta)$ for control. Then our mask can be updated as:
\begin{equation}
    M'_{h, w, t}(\theta) = 
\begin{cases} 
f_{entity}(\theta) & \text{if } t \in r_{M_{h,w}} \\
f_{other}(\theta) & \text{if } t \in R_O \\
0 & \text{otherwise} 
\end{cases}
\end{equation}
This leads to our final attention control function in the cross-attention module:
\begin{equation}
    A''_{h, w, t} = A_{h, w, t} \times M'_{h, w, t}(\theta)
\end{equation}
Here, $\theta$ represents the current time step of the diffusion process. For detailed configuration of $f_{entity}(\theta), f_{other}(\theta)$, please refer to our supplementary materials.

With this dynamic reweighting method, we guide the model to focus on the primary information in different stages as shown in Figure \ref{cross_comp_weight}. By applying phase-wise control to entities and other information, our method enables the attention map to distinguish between entities and image backgrounds more effectively. This is particularly evident in the outline parts of the female in Figure \ref{cross_our}, during the third and fourth periods.

\section{Experiments}
\label{exp}
Our approach is a training-free method, utilizing the COCO2014 validation set~\cite{lin2014microsoft} for evaluation purposes. We adopt the latest Stable Diffusion XL 1.0~\cite{podell2023sdxl} as our baseline and compare it with Structured Diffusion~\cite{feng2022training}. 
In Section \ref{41}, the qualitative analysis is performed. Subsequently, the quantitative findings of FID~\cite{heusel2017gans}, CLIP Score~\cite{hessel2021clipscore} and ImageReward~\cite{xu2023imagereward} are analysed in Section \ref{42}. We also conduct a self-designed semi-human evaluation on specific alignment tasks to examine our model's alignment capability.
In Section \ref{43}, we conduct ablation study to examine the distinct roles and impacts of each component employed in our approach.

\begin{figure*}[!t]
\centering
\rotatebox[origin=c]{90}{Stable Diffusion XL 1.0}\quad 
\begin{minipage}{.22\textwidth}
  \centering
  \parbox[c][2.5em][t]{\linewidth}{
  \fontsize{8pt}{8pt}\selectfont\textcolor{red}{Two} giraffes crossing paths on a green and grassy field.}\\ 
  \begin{subfigure}{.48\linewidth}
    \centering
    \includegraphics[width=\linewidth]{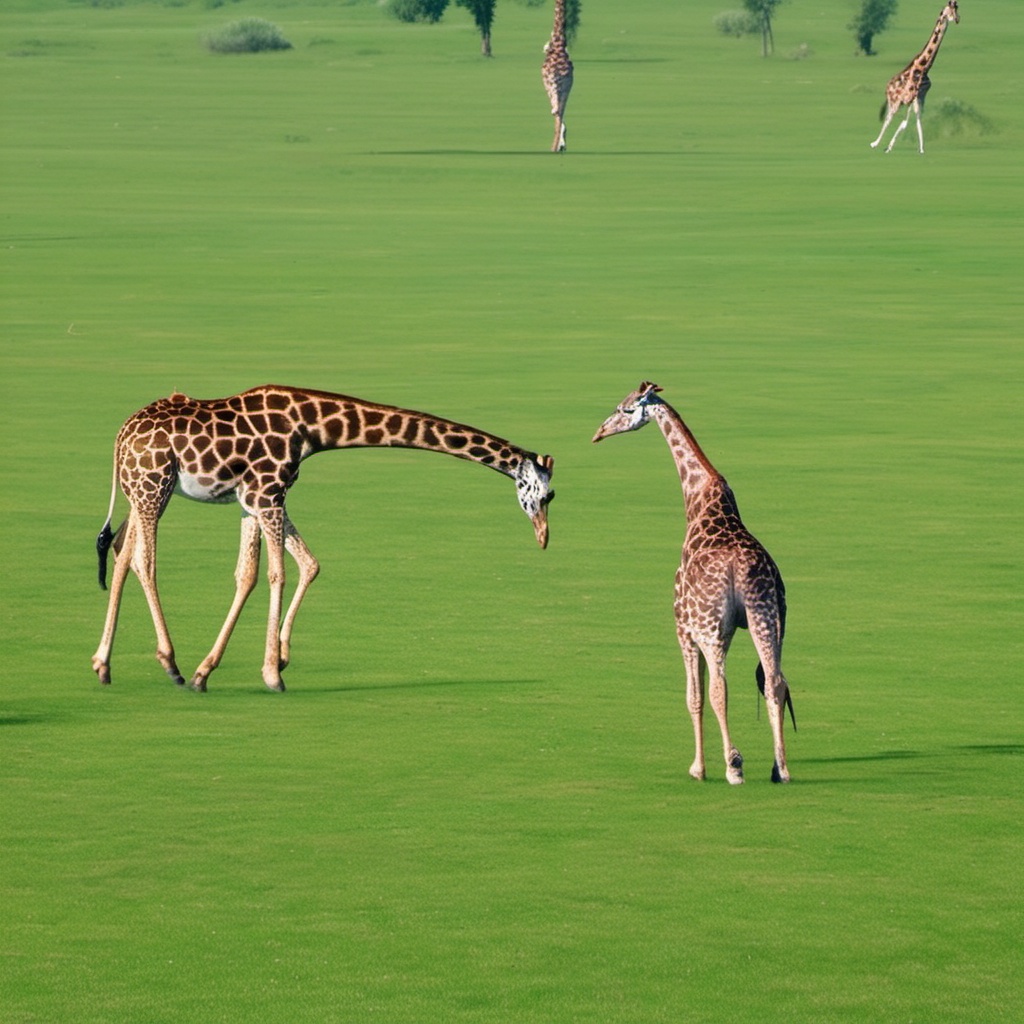}
    
  \end{subfigure}%
  \hfill 
  \begin{subfigure}{.48\linewidth}
    \centering
    \includegraphics[width=\linewidth]{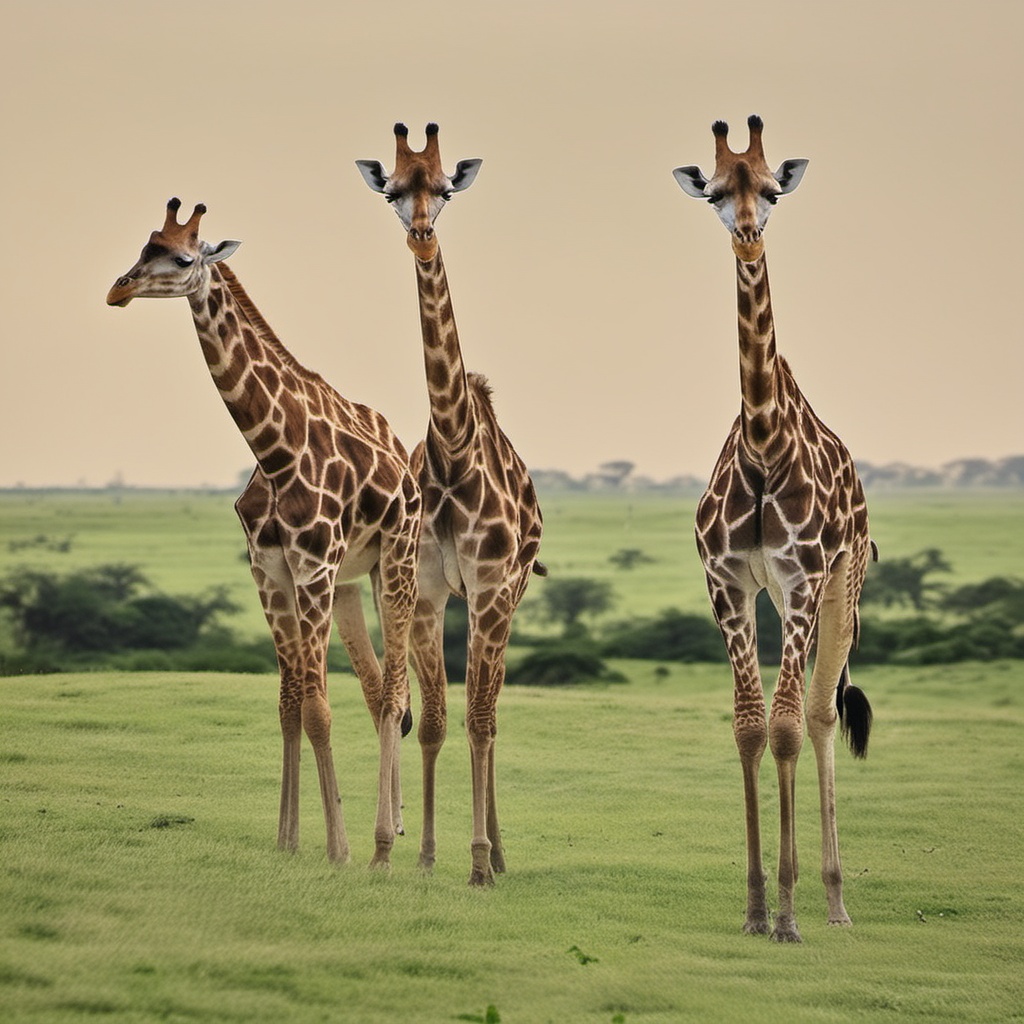}
    
  \end{subfigure}\\
  \begin{subfigure}{.48\linewidth}
    \centering
    \includegraphics[width=\linewidth]{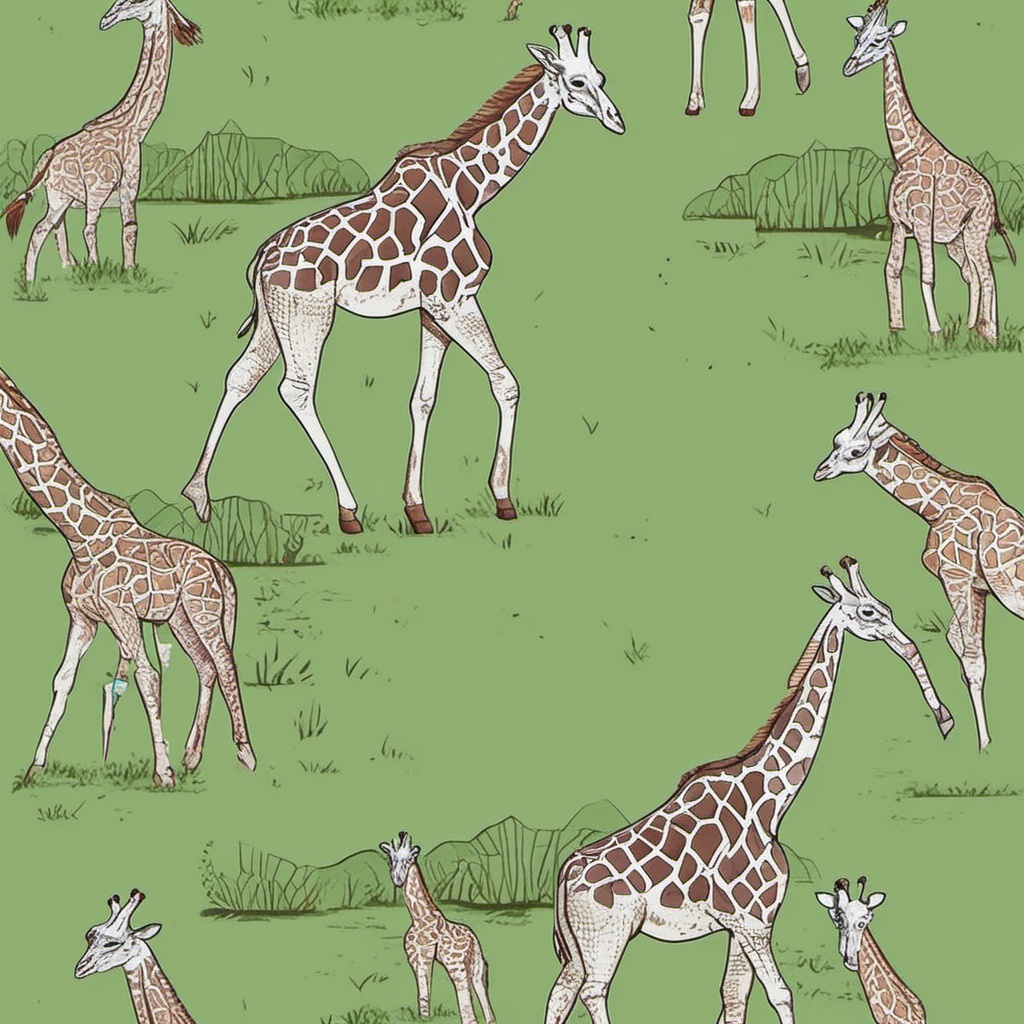}
    
  \end{subfigure}%
  \hfill 
  \begin{subfigure}{.48\linewidth}
    \centering
    \includegraphics[width=\linewidth]{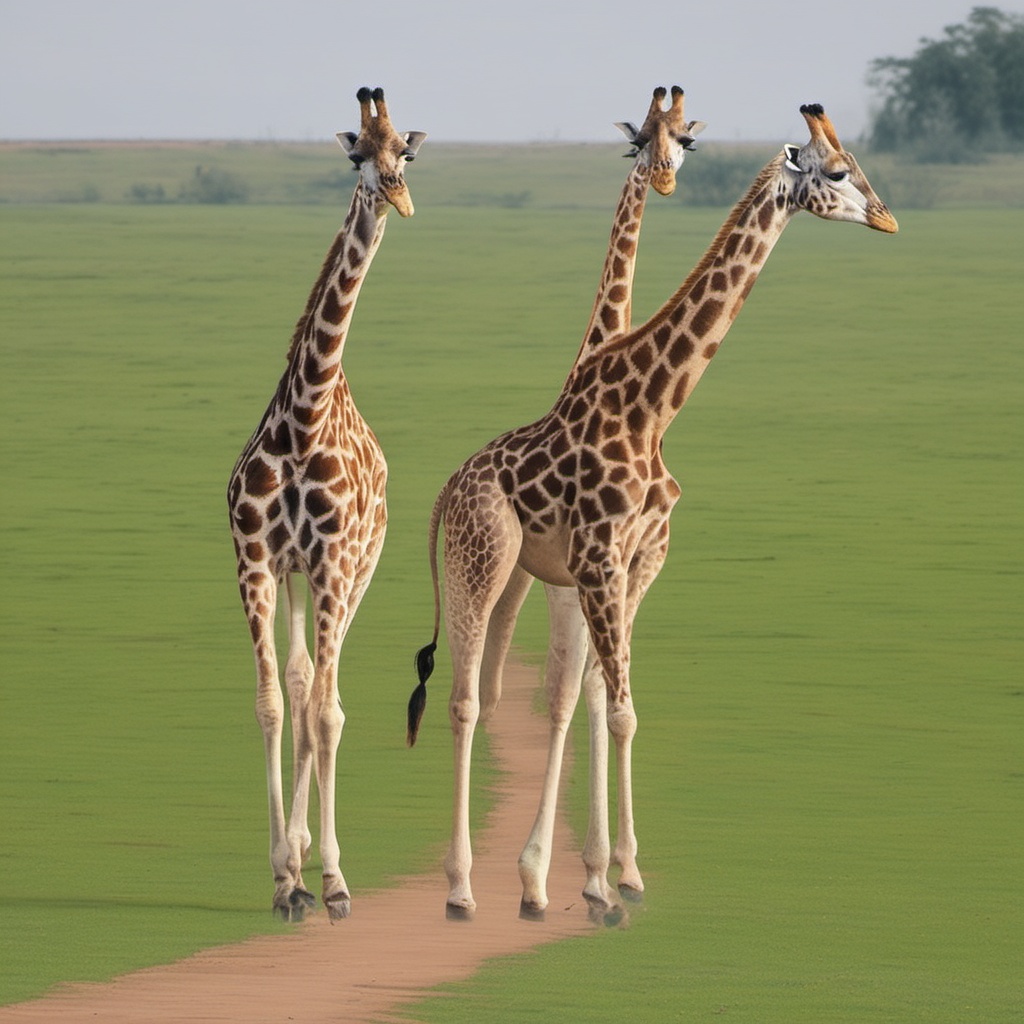}
    
  \end{subfigure}\\
  (a)
\end{minipage}%
\hfill 
\begin{minipage}{.22\textwidth}
  \centering
  \parbox[c][2.5em][t]{\linewidth}{
  \fontsize{8pt}{8pt}\selectfont\textcolor{red}{A boy} in front of \textcolor{blue}{a female}.}\\ 
  \begin{subfigure}{.48\linewidth}
    \centering
    \includegraphics[width=\linewidth]{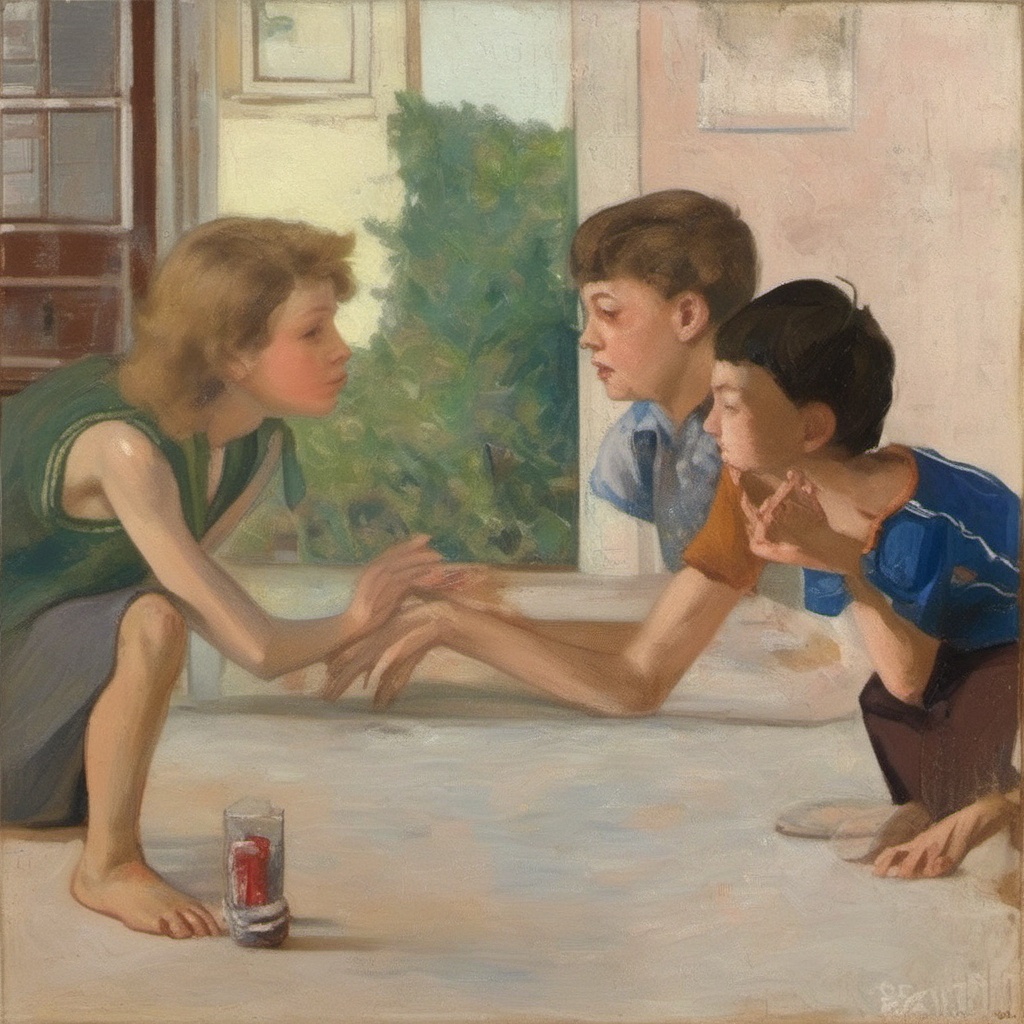}
    
  \end{subfigure}%
  \hfill 
  \begin{subfigure}{.48\linewidth}
    \centering
    \includegraphics[width=\linewidth]{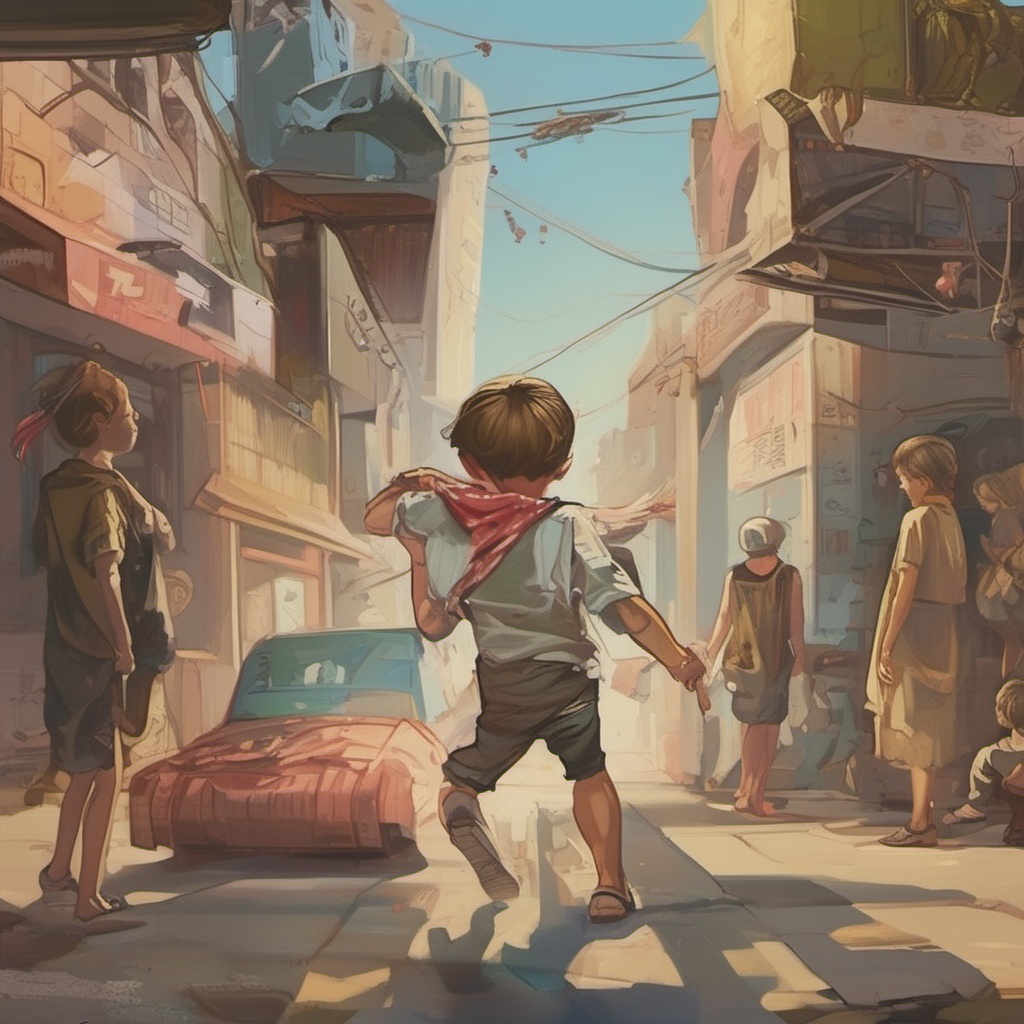}
    
  \end{subfigure}\\
  \begin{subfigure}{.48\linewidth}
    \centering
    \includegraphics[width=\linewidth]{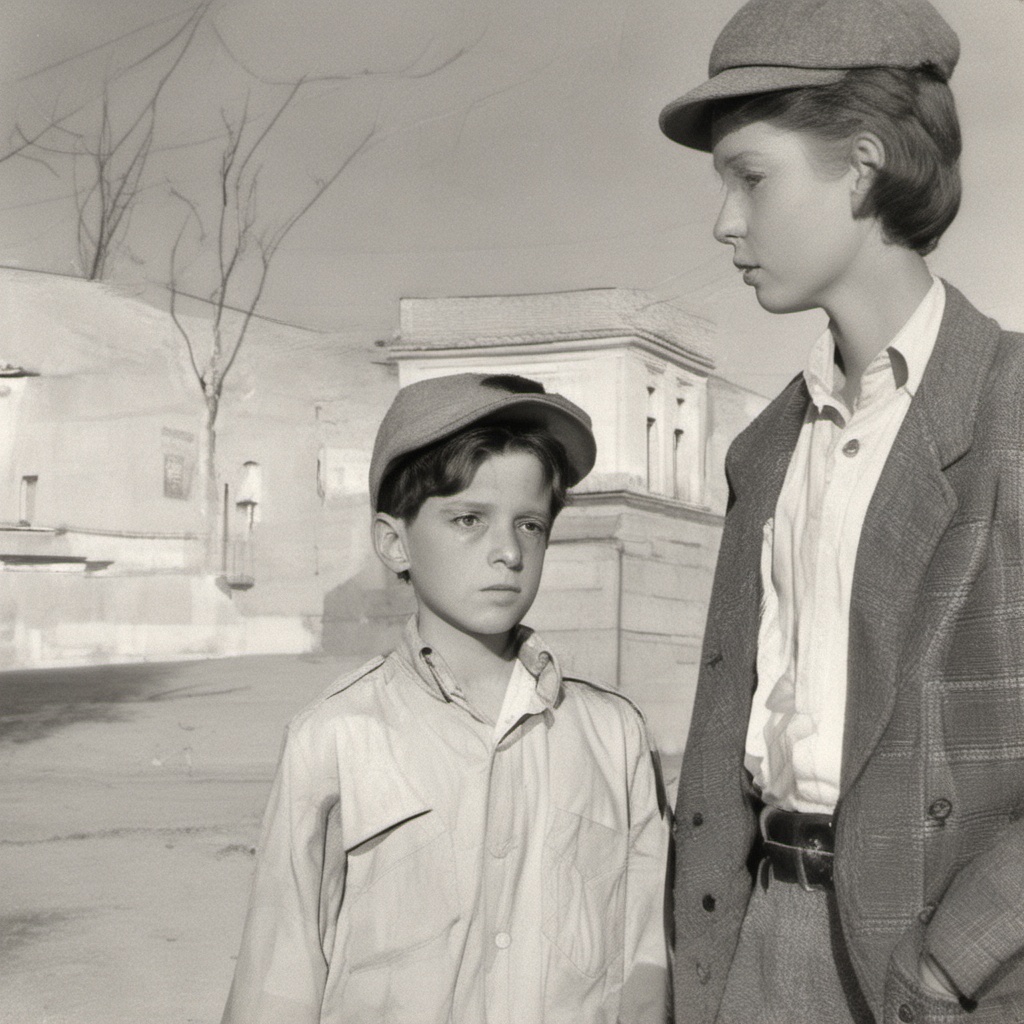}
    
  \end{subfigure}%
  \hfill 
  \begin{subfigure}{.48\linewidth}
    \centering
    \includegraphics[width=\linewidth]{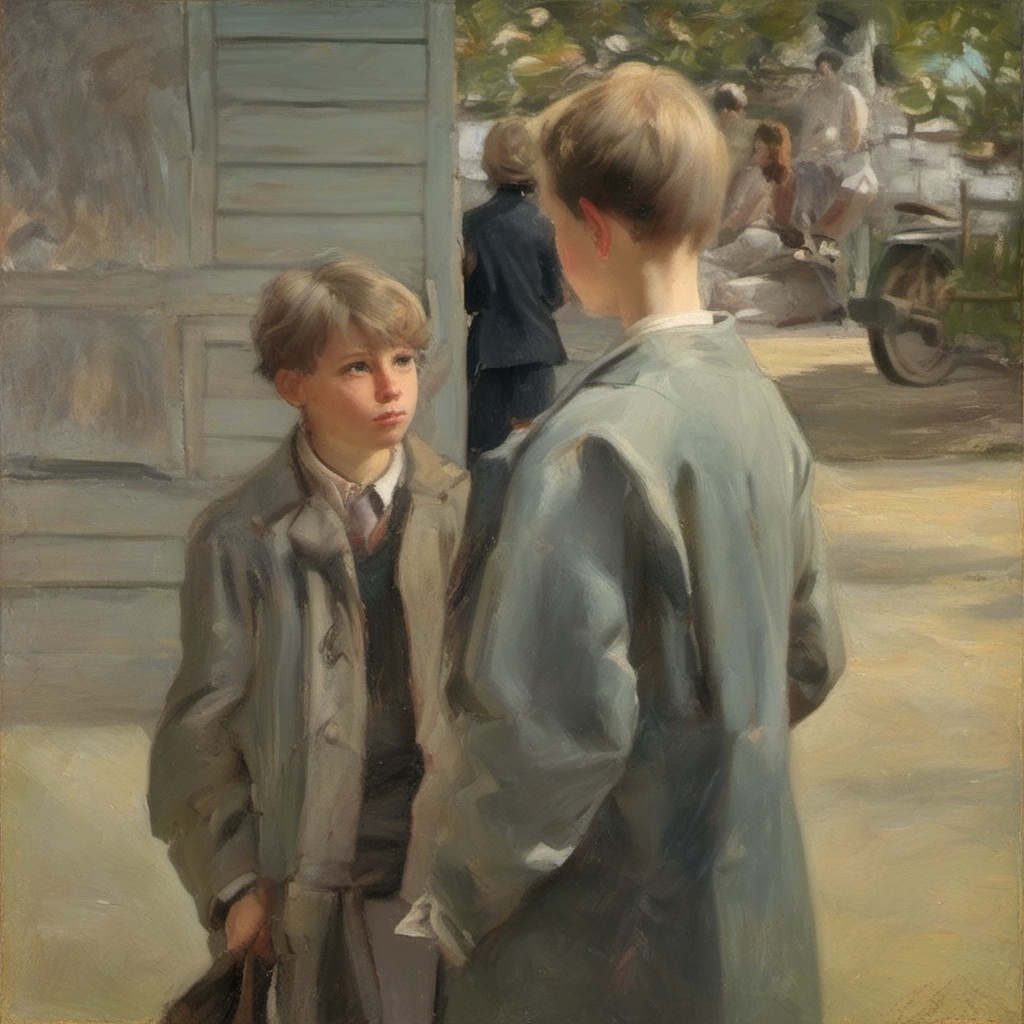}
    
  \end{subfigure}
    (b)
\end{minipage}%
\hfill
\begin{minipage}{.22\textwidth}
  \centering
  \parbox[c][2.5em][t]{\linewidth}{
  \fontsize{8pt}{8pt}\selectfont\textcolor{red}{A giraffe} and \textcolor{blue}{zebra} inside of a zoo enclosure.}\\ 
  \begin{subfigure}{.48\linewidth}
    \centering
    \includegraphics[width=\linewidth]{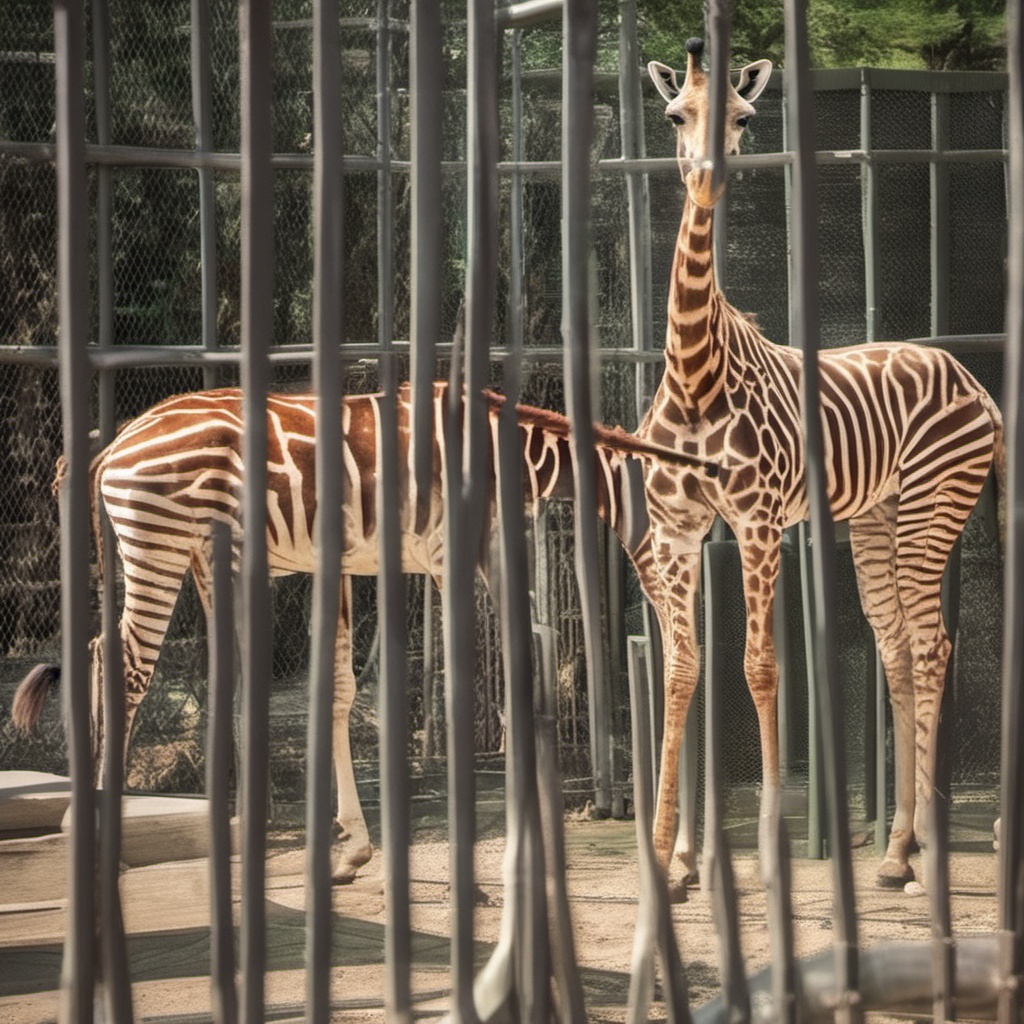}
    
  \end{subfigure}%
  \hfill 
  \begin{subfigure}{.48\linewidth}
    \centering
    \includegraphics[width=\linewidth]{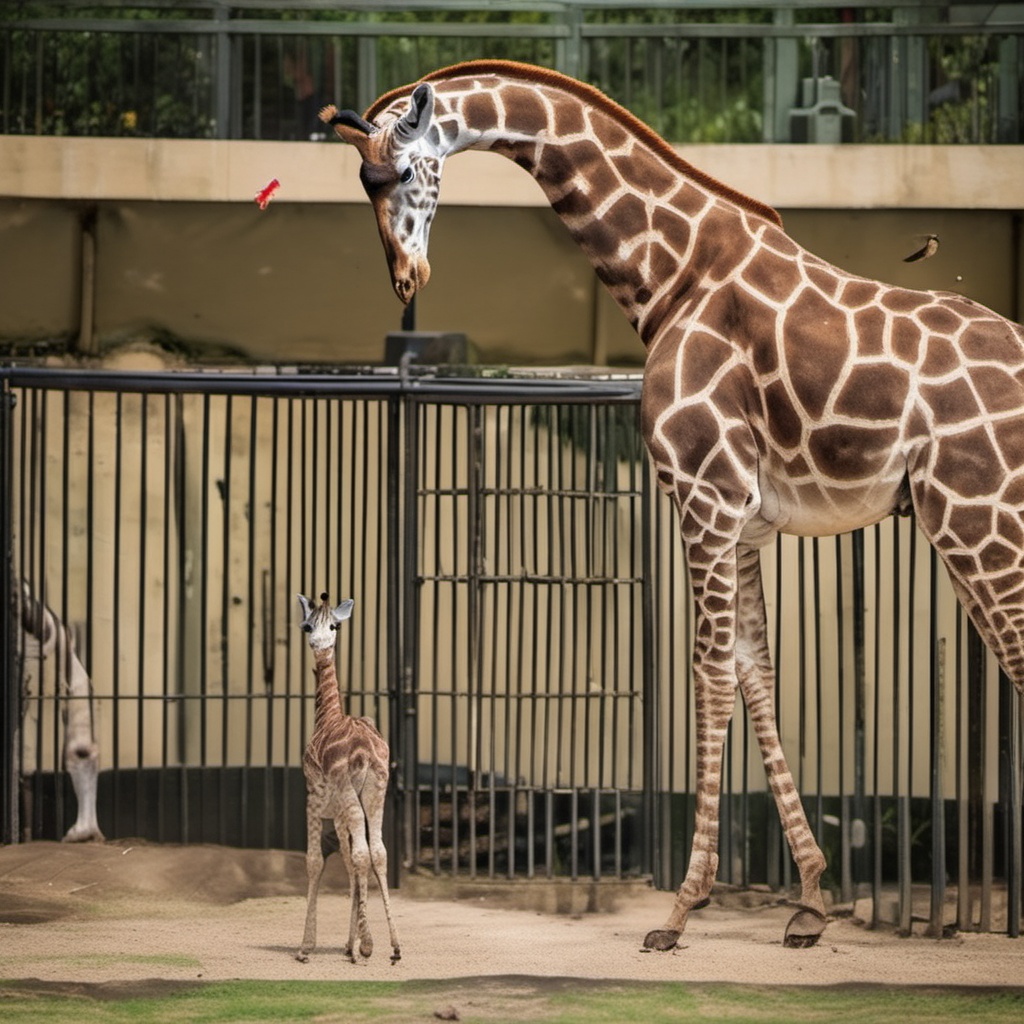}
    
  \end{subfigure}\\
  \begin{subfigure}{.48\linewidth}
    \centering
    \includegraphics[width=\linewidth]{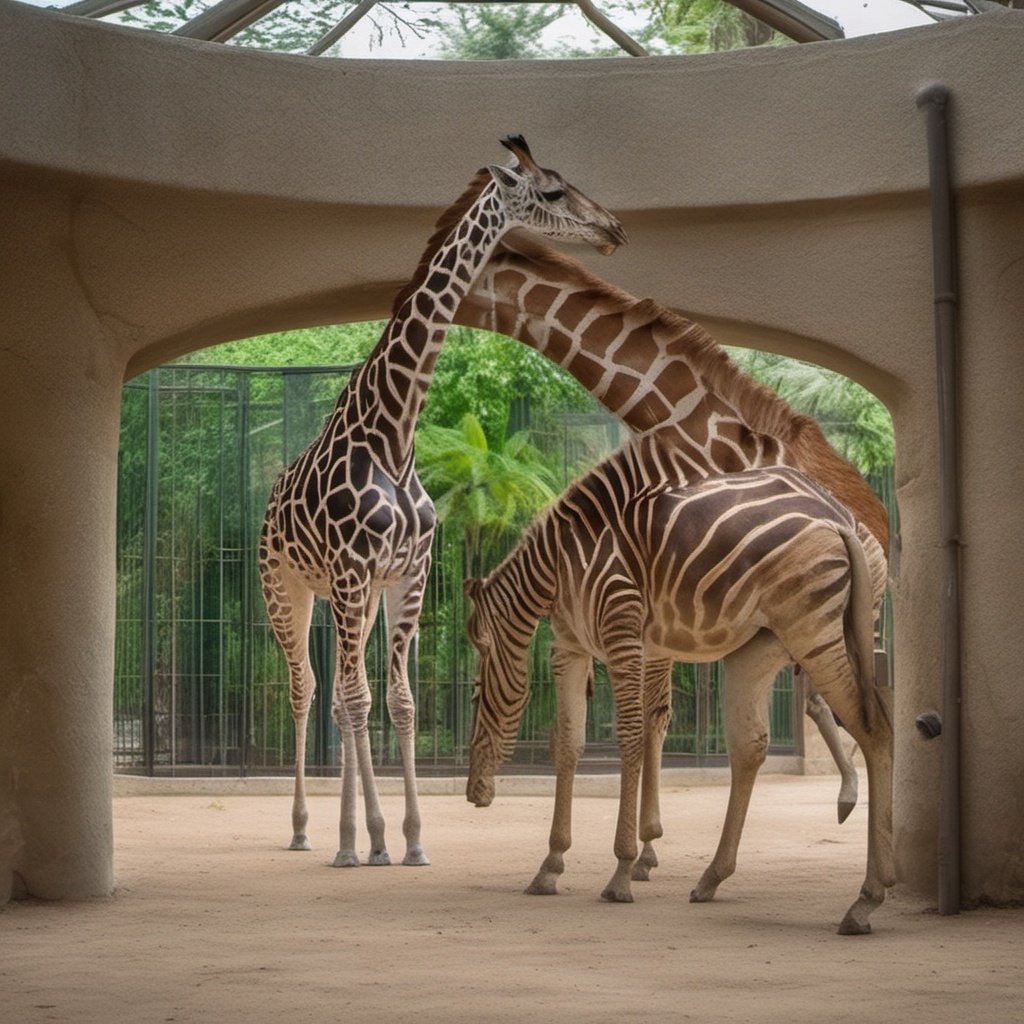}
    
  \end{subfigure}%
  \hfill 
  \begin{subfigure}{.48\linewidth}
    \centering
    \includegraphics[width=\linewidth]{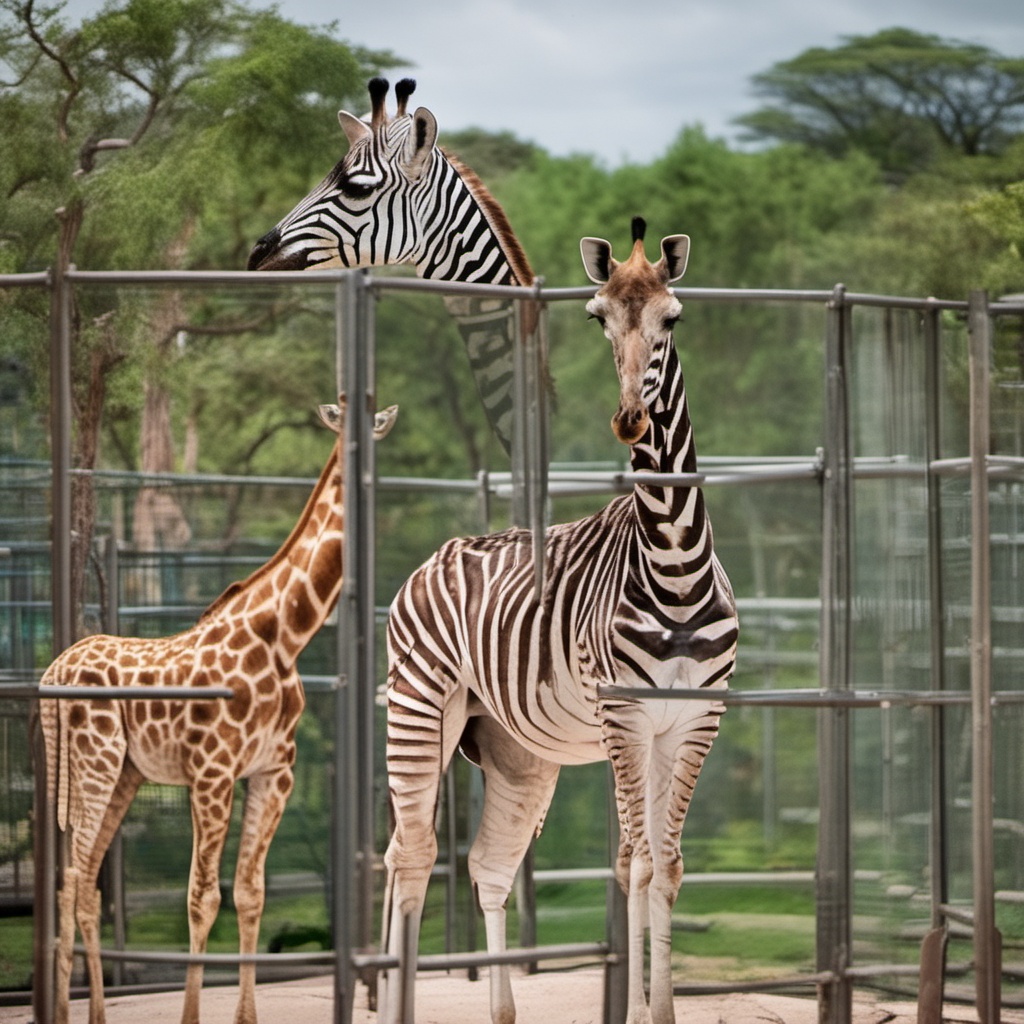}
    
  \end{subfigure}
    (c)
\end{minipage}%
\hfill
\begin{minipage}{.22\textwidth}
  \centering
  \parbox[c][2.5em][t]{\linewidth}{
  \fontsize{8pt}{8pt}\selectfont\textcolor{red}{Three birds}  perched on a gutter on a building.}\\ 
  \begin{subfigure}{.48\linewidth}
    \centering
    \includegraphics[width=\linewidth]{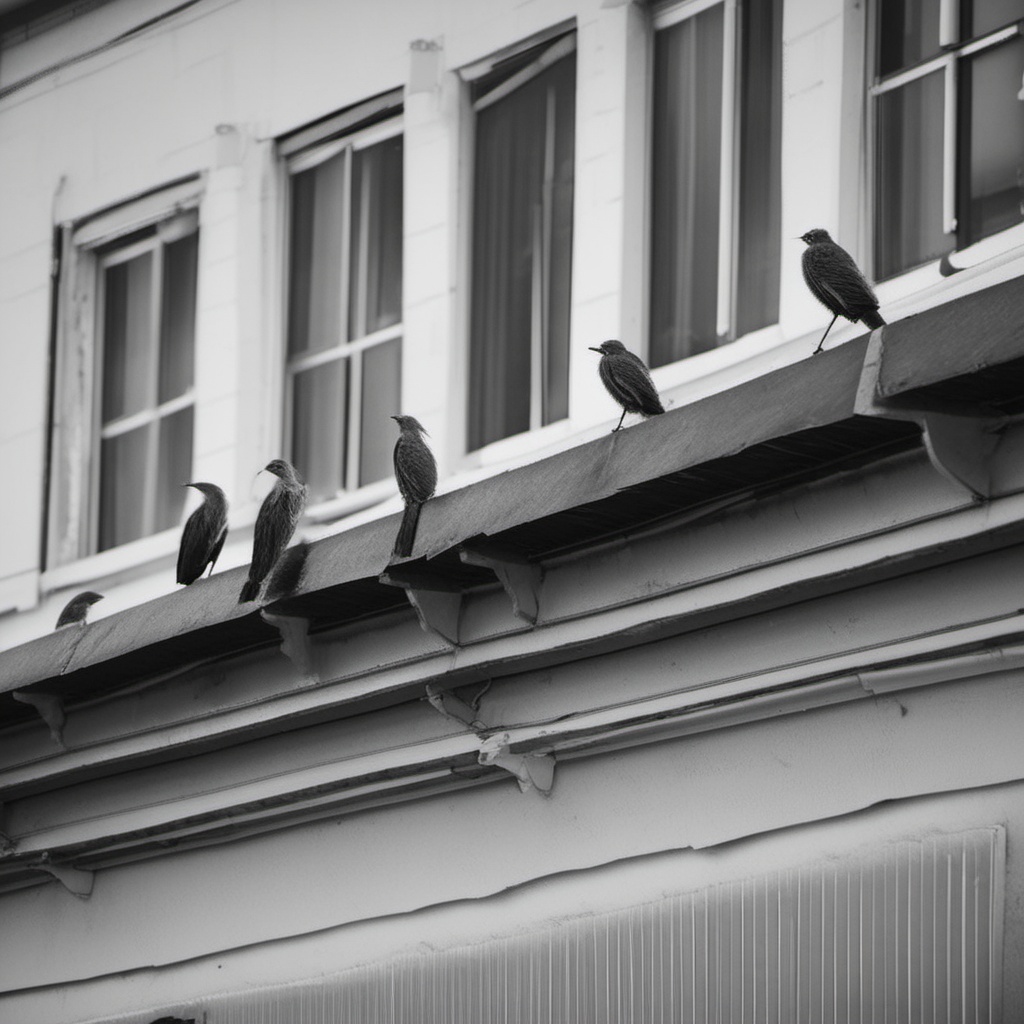}
    
  \end{subfigure}%
  \hfill 
  \begin{subfigure}{.48\linewidth}
    \centering
    \includegraphics[width=\linewidth]{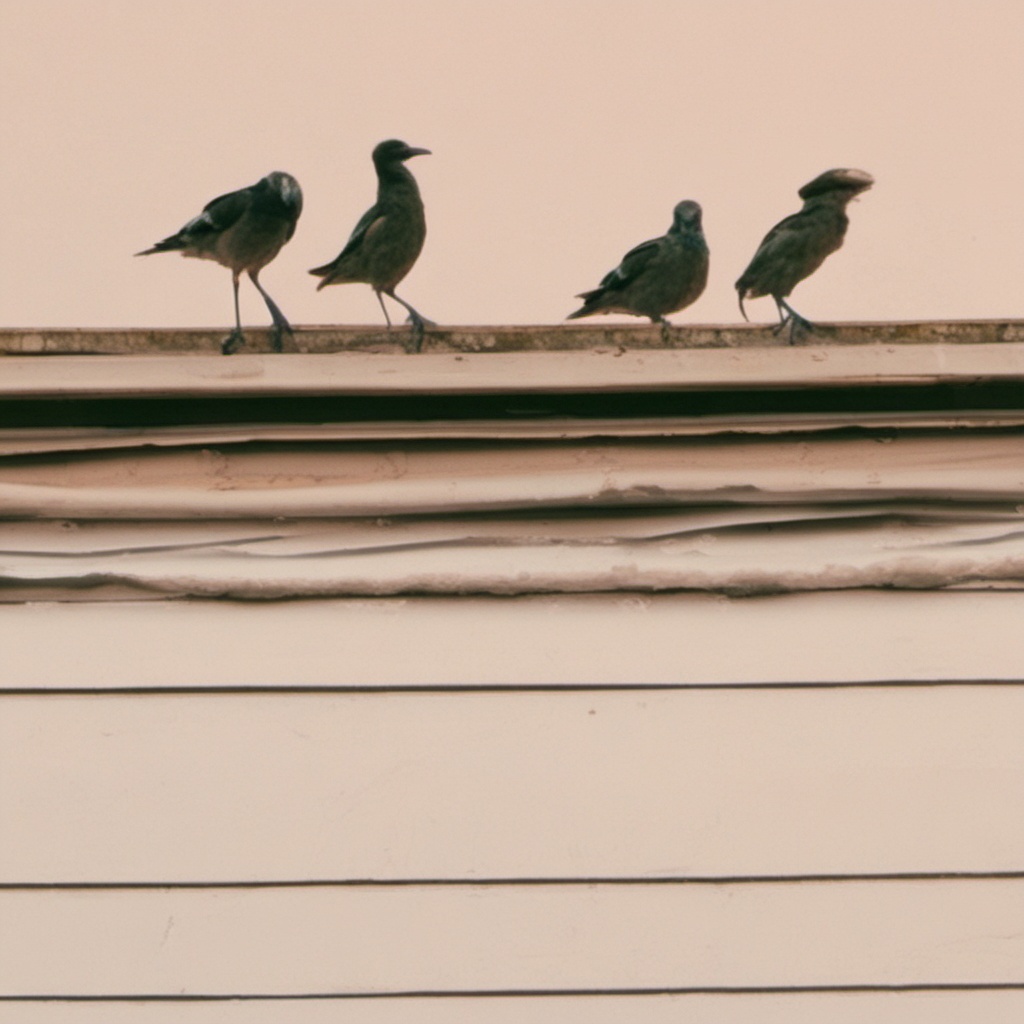}
    
  \end{subfigure}\\
  \begin{subfigure}{.48\linewidth}
    \centering
    \includegraphics[width=\linewidth]{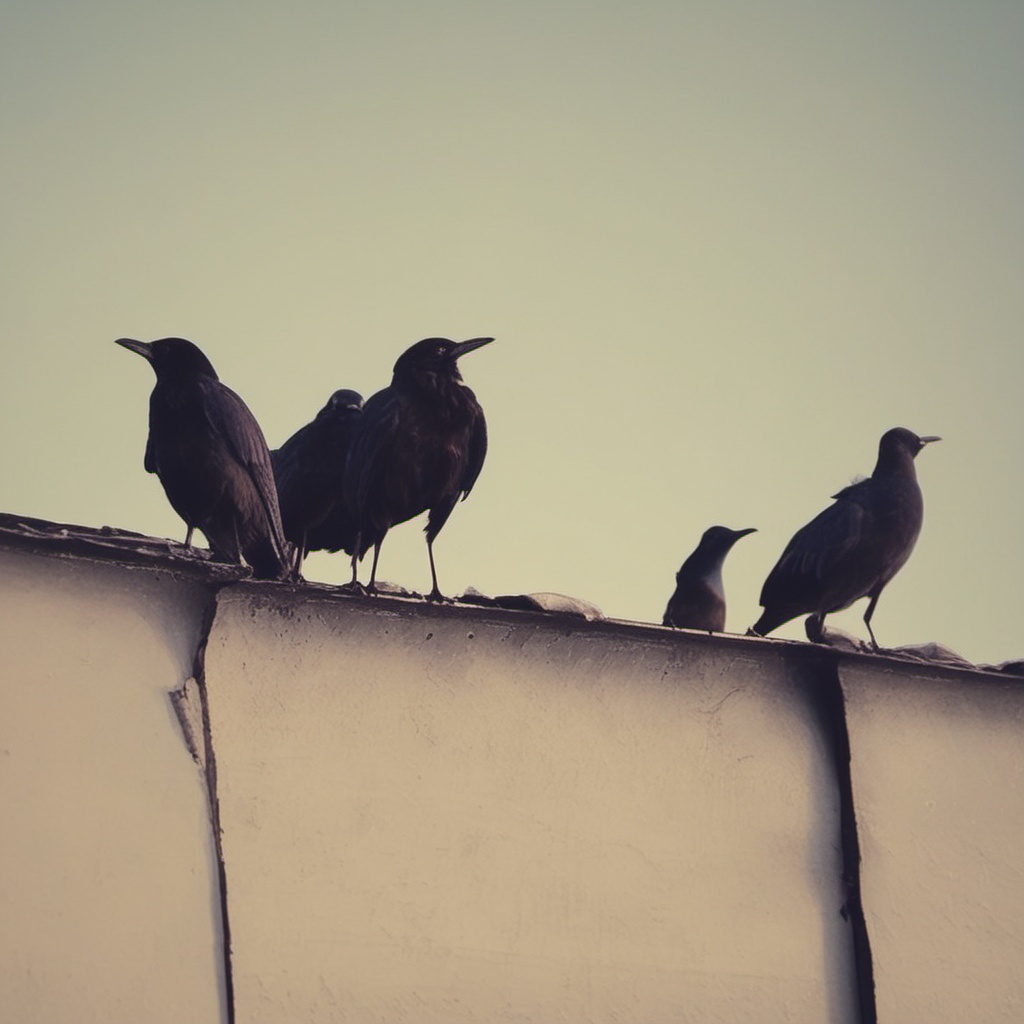}
    
  \end{subfigure}%
  \hfill 
  \begin{subfigure}{.48\linewidth}
    \centering
    \includegraphics[width=\linewidth]{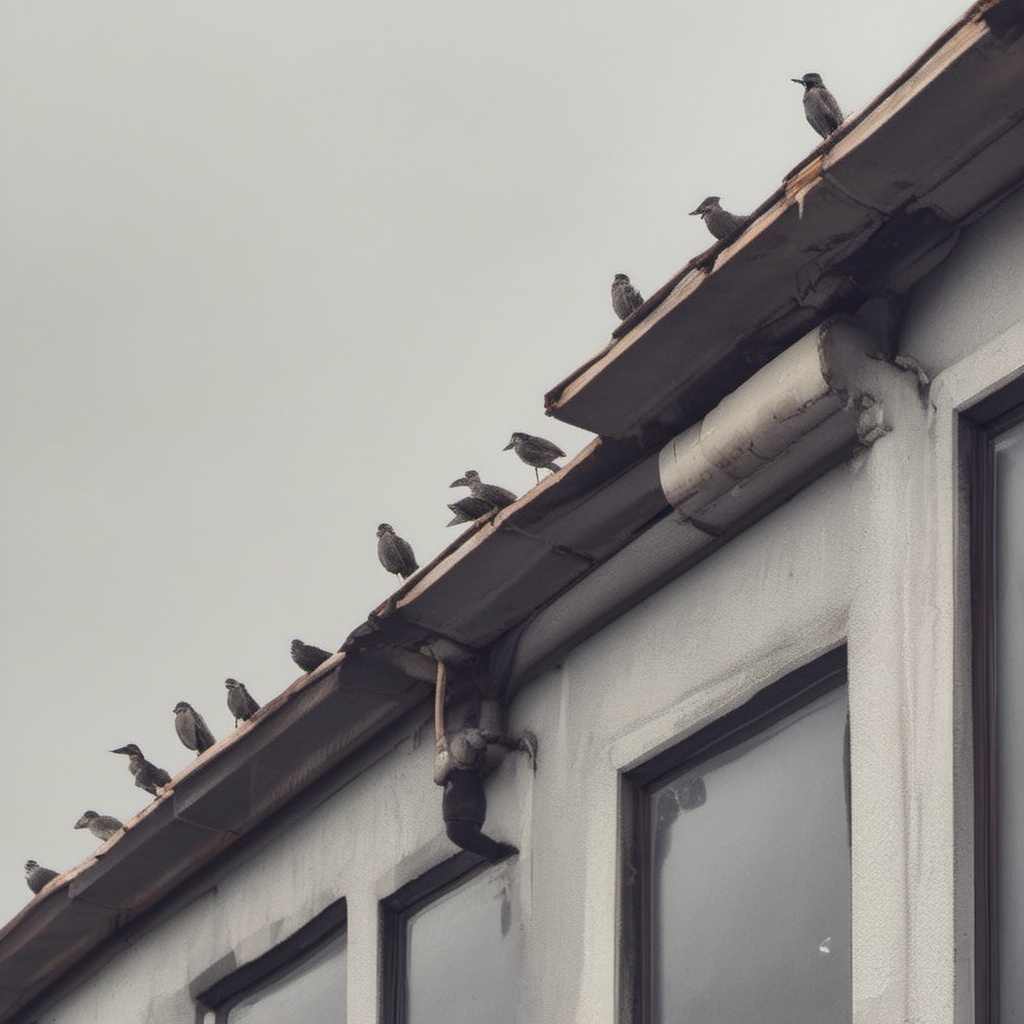}
    
  \end{subfigure}
    (d)
\end{minipage}%

\vspace{0.1em} 

\rotatebox[origin=c]{90}{Structured Diffusion}\quad 
\begin{minipage}{.22\textwidth}
  \centering
  \begin{subfigure}{.48\linewidth}
    \centering
    \includegraphics[width=\linewidth]{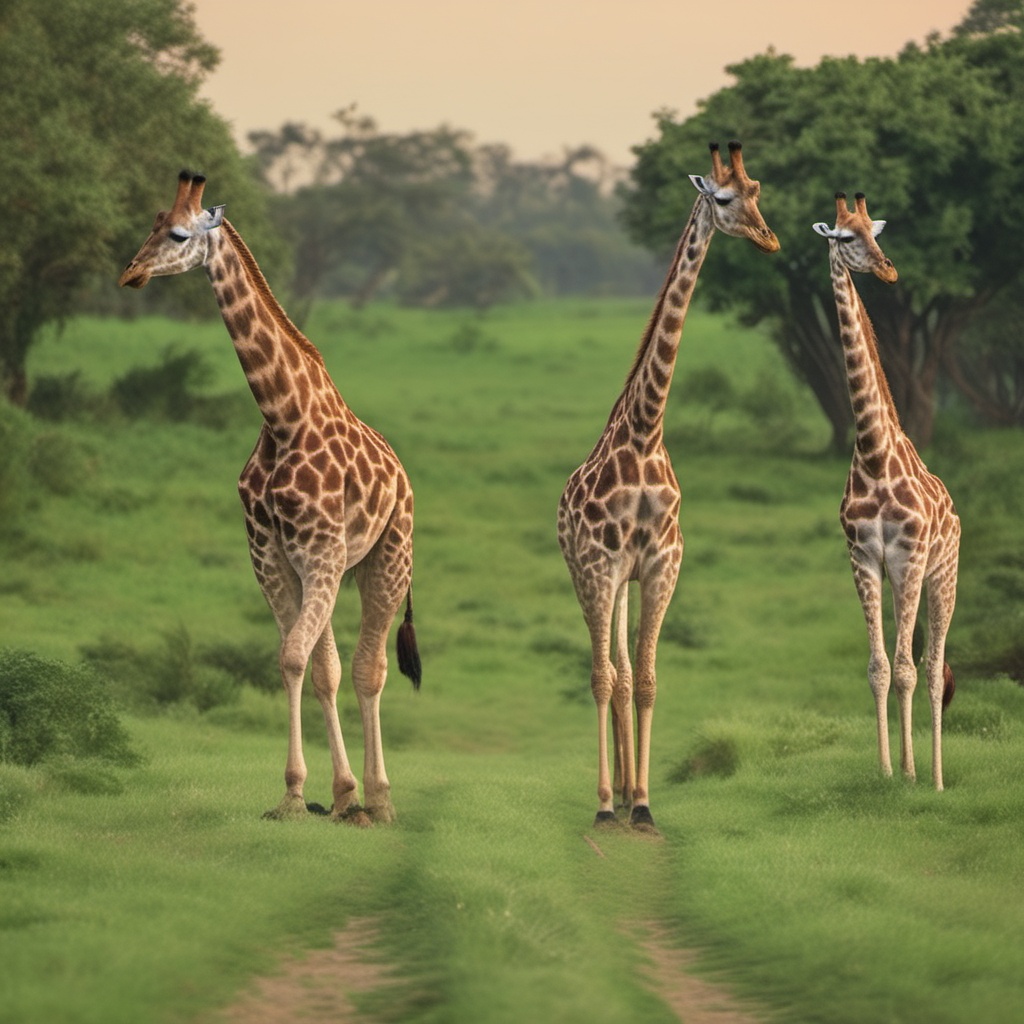}
    
  \end{subfigure}%
  \hfill 
  \begin{subfigure}{.48\linewidth}
    \centering
    \includegraphics[width=\linewidth]{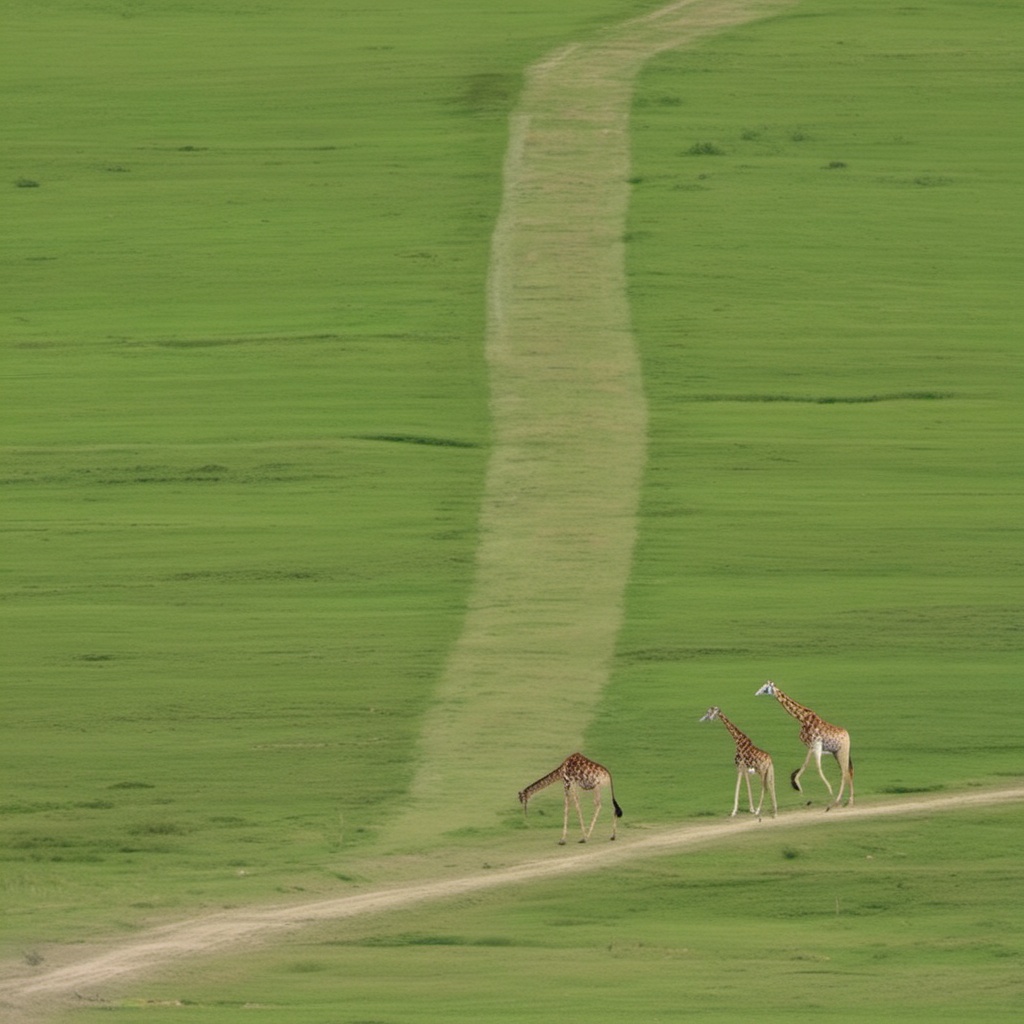}
    
  \end{subfigure}\\
  \begin{subfigure}{.48\linewidth}
    \centering
    \includegraphics[width=\linewidth]{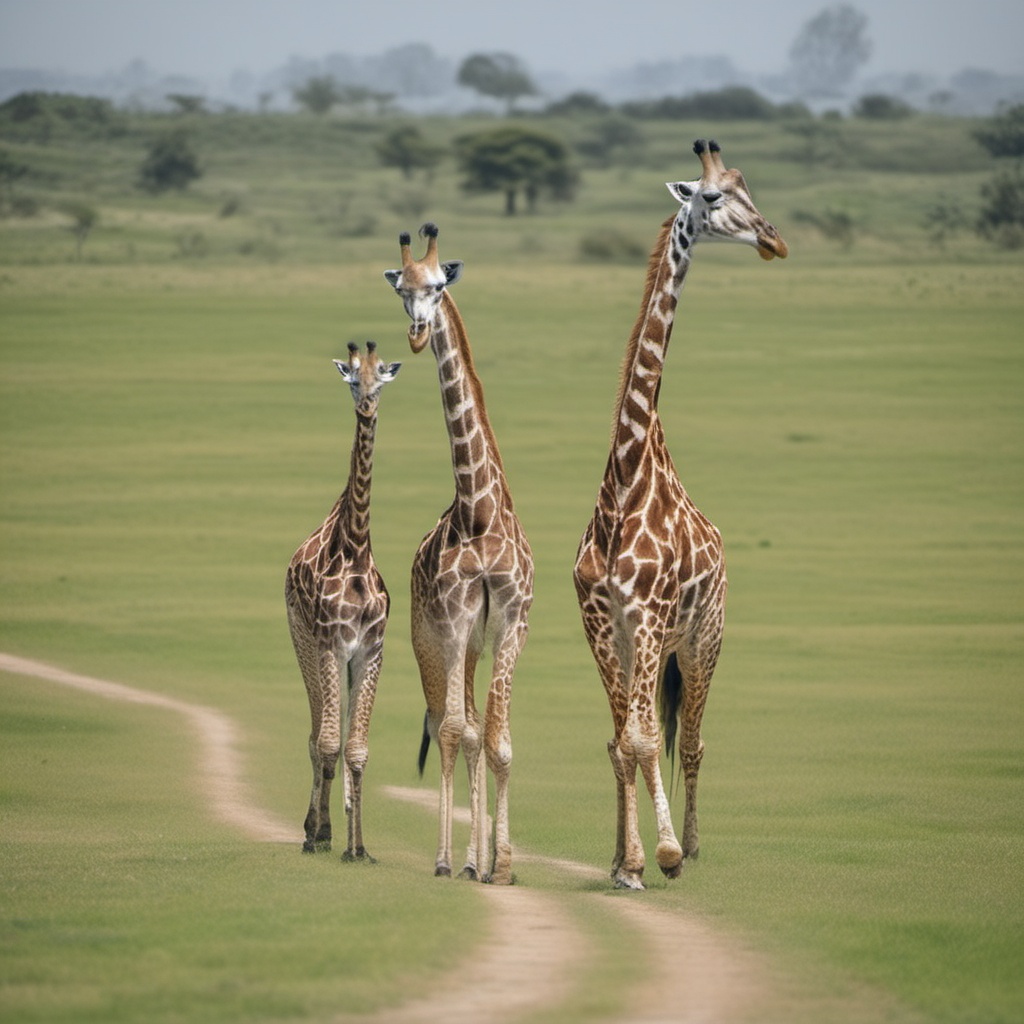}
    
  \end{subfigure}%
  \hfill 
  \begin{subfigure}{.48\linewidth}
    \centering
    \includegraphics[width=\linewidth]{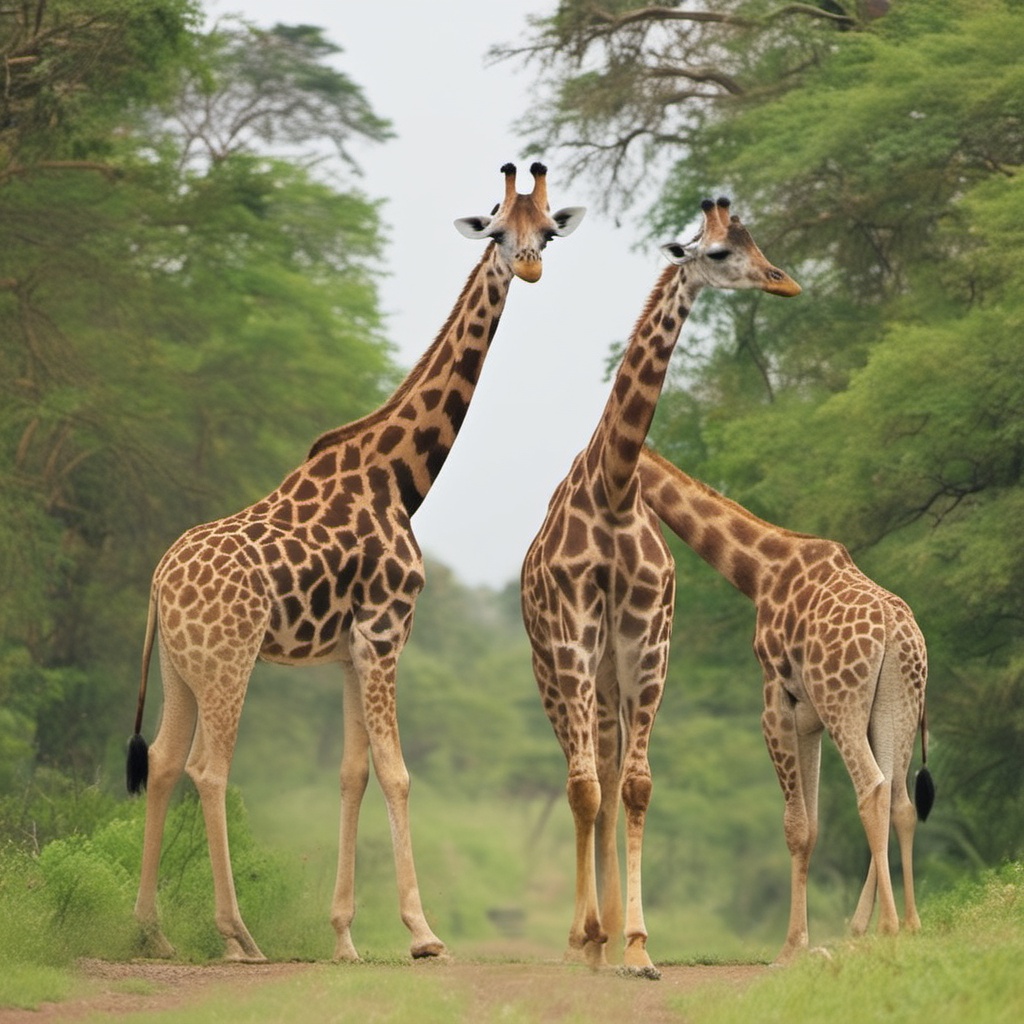}
    
  \end{subfigure}
    (e)
\end{minipage}%
\hfill 
\begin{minipage}{.22\textwidth}
  \centering
  \begin{subfigure}{.48\linewidth}
    \centering
    \includegraphics[width=\linewidth]{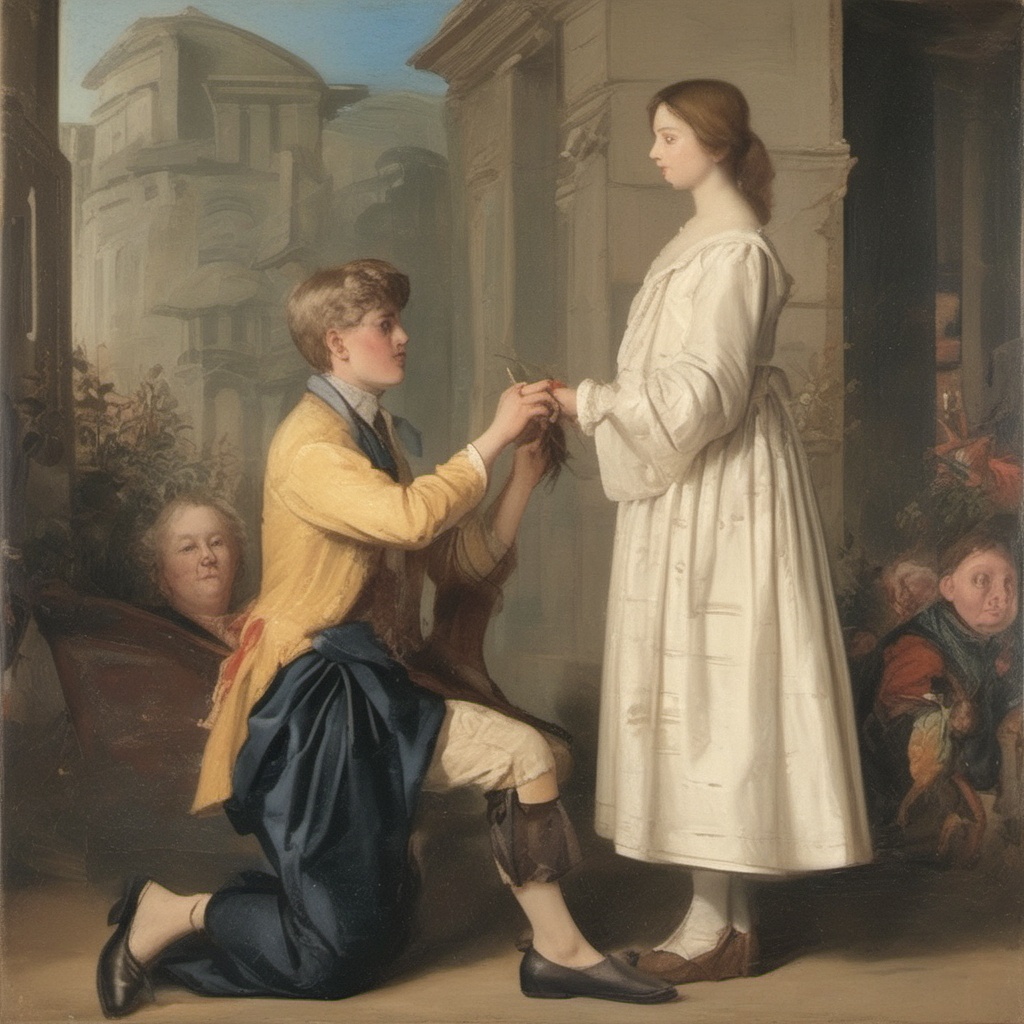}
    
  \end{subfigure}%
  \hfill 
  \begin{subfigure}{.48\linewidth}
    \centering
    \includegraphics[width=\linewidth]{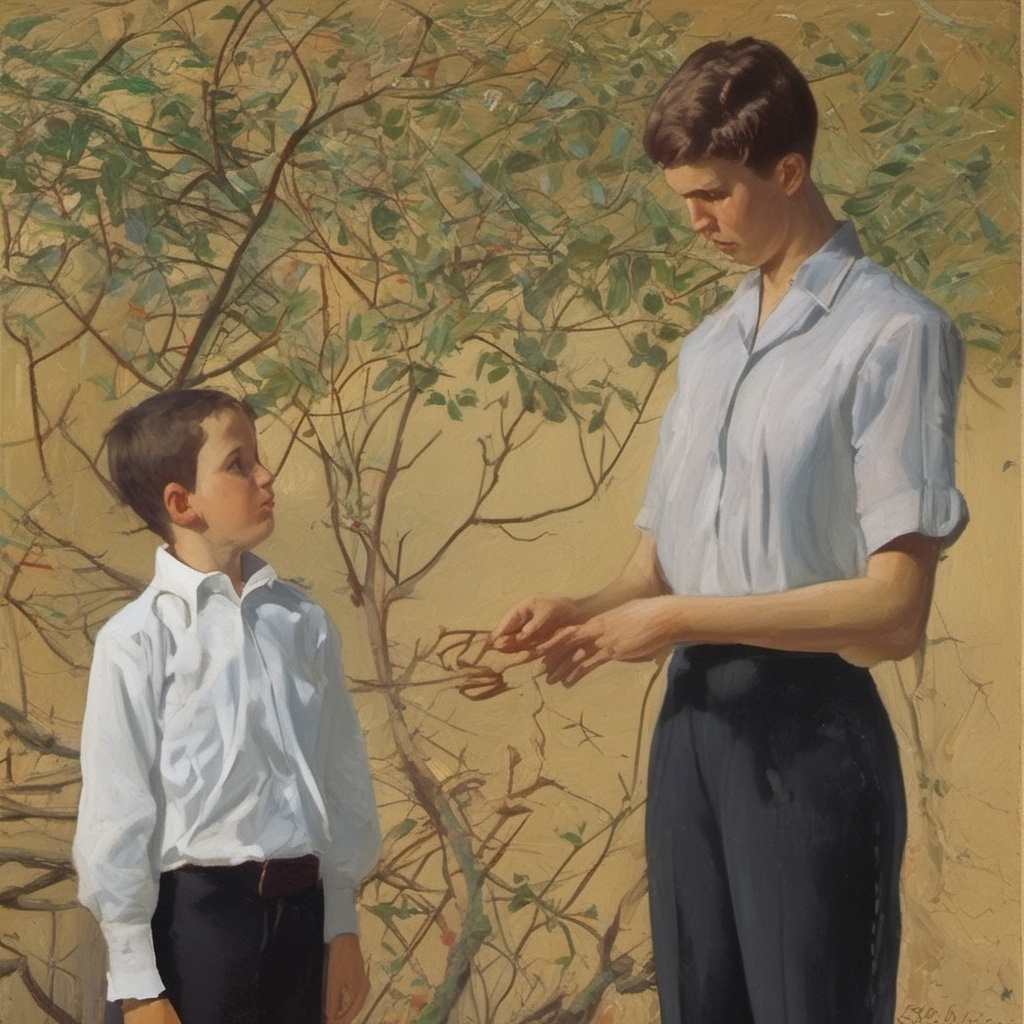}
    
  \end{subfigure}\\
  \begin{subfigure}{.48\linewidth}
    \centering
    \includegraphics[width=\linewidth]{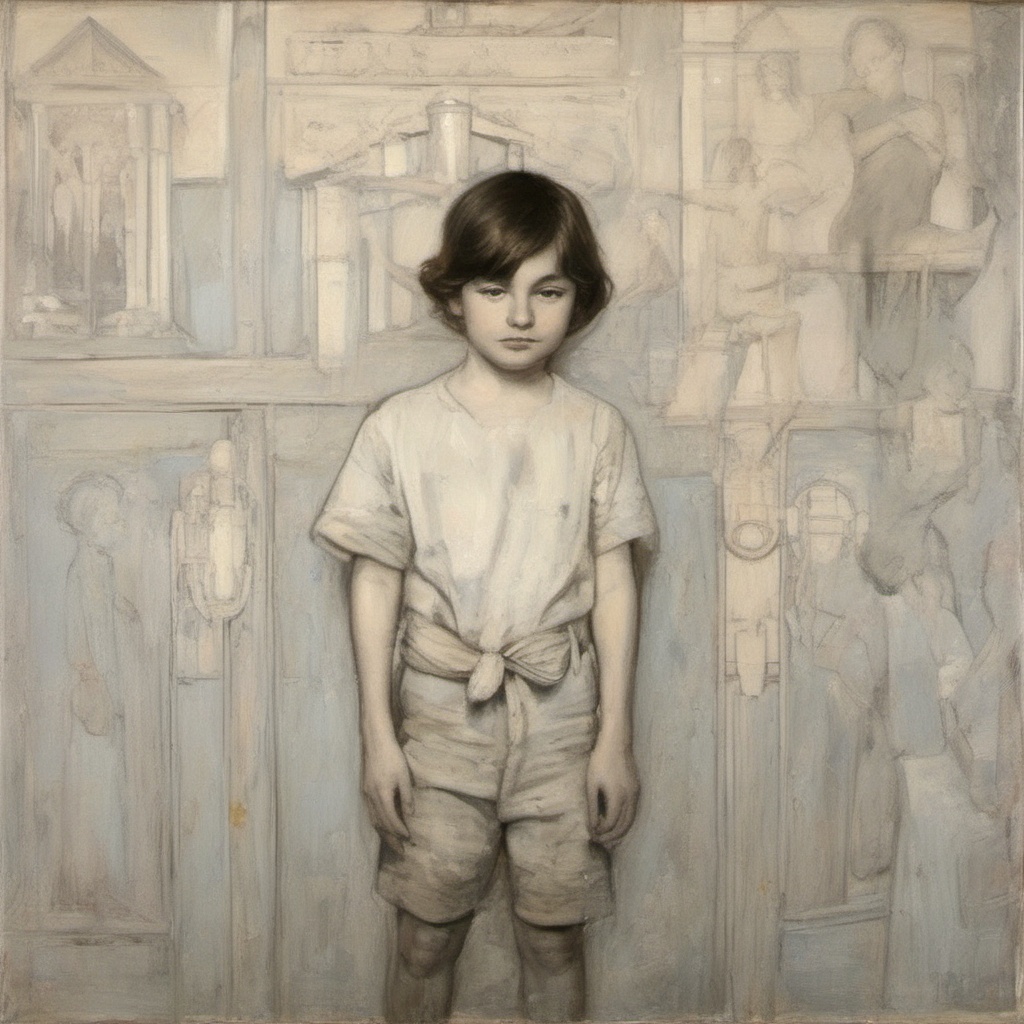}
    
  \end{subfigure}%
  \hfill 
  \begin{subfigure}{.48\linewidth}
    \centering
    \includegraphics[width=\linewidth]{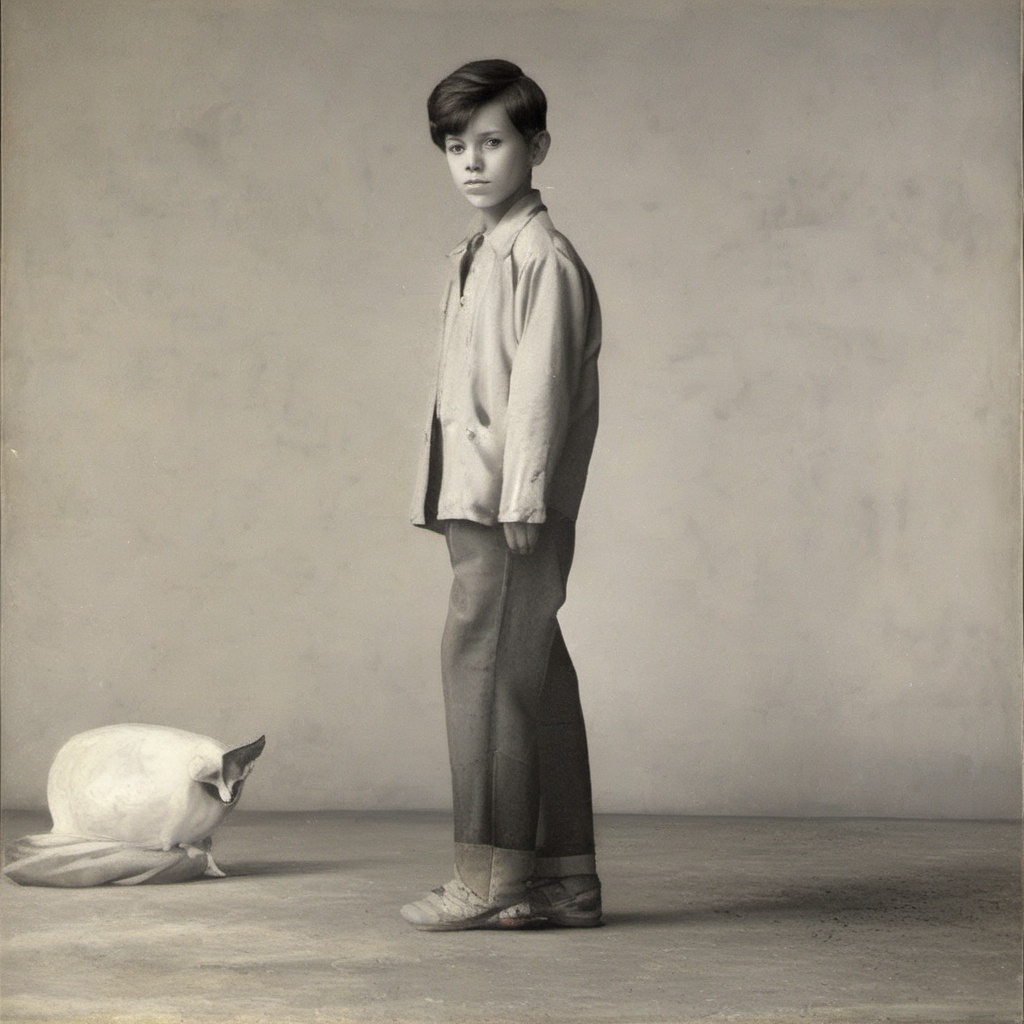}
    
  \end{subfigure}
    (f)
\end{minipage}%
\hfill
\begin{minipage}{.22\textwidth}
  \centering
  \begin{subfigure}{.48\linewidth}
    \centering
    \includegraphics[width=\linewidth]{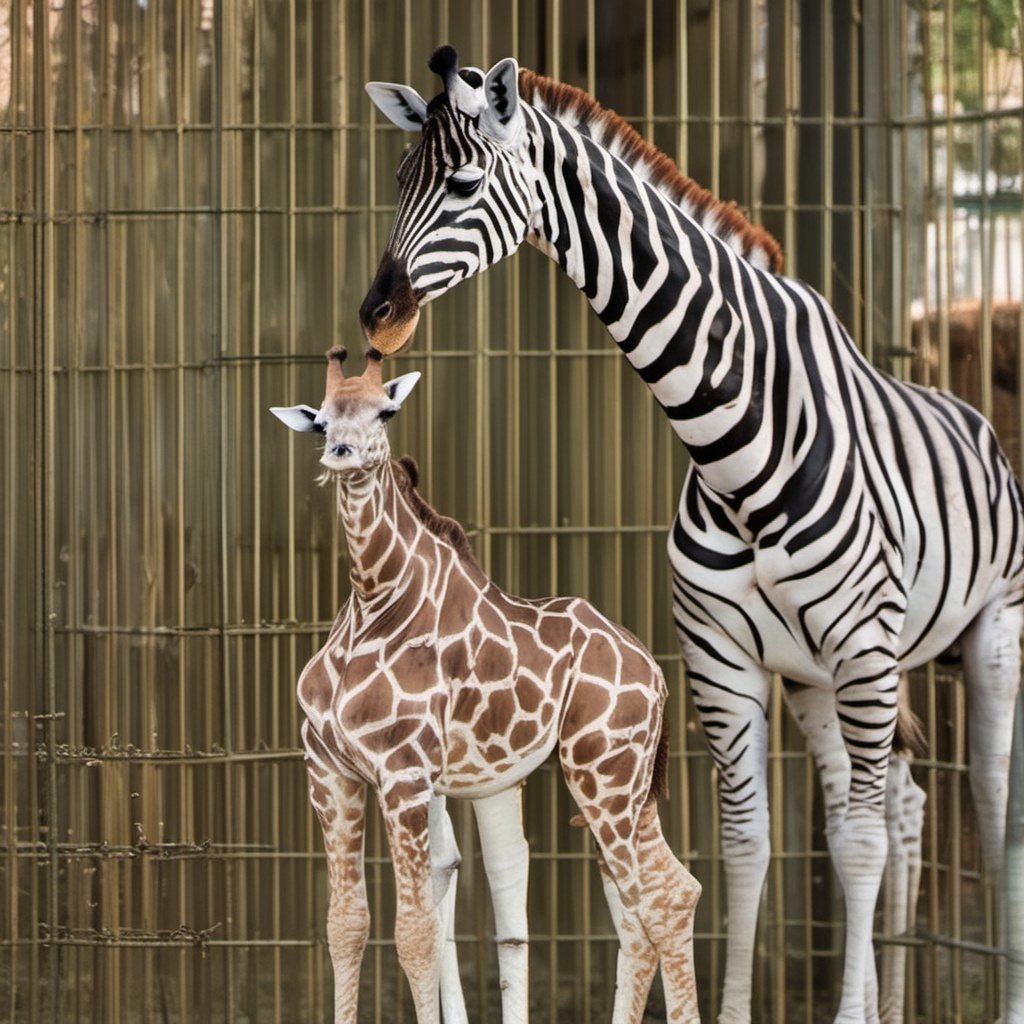}
    
  \end{subfigure}%
  \hfill 
  \begin{subfigure}{.48\linewidth}
    \centering
    \includegraphics[width=\linewidth]{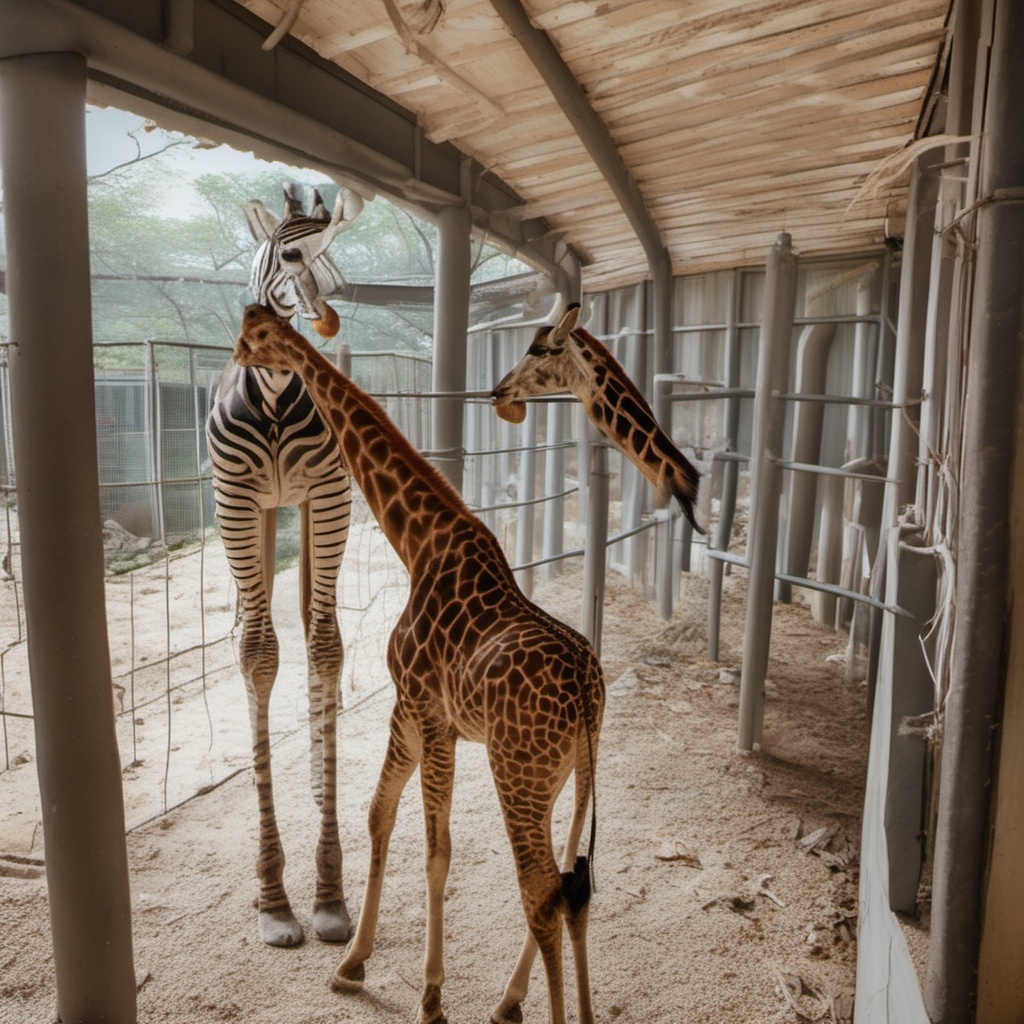}
    
  \end{subfigure}\\
  \begin{subfigure}{.48\linewidth}
    \centering
    \includegraphics[width=\linewidth]{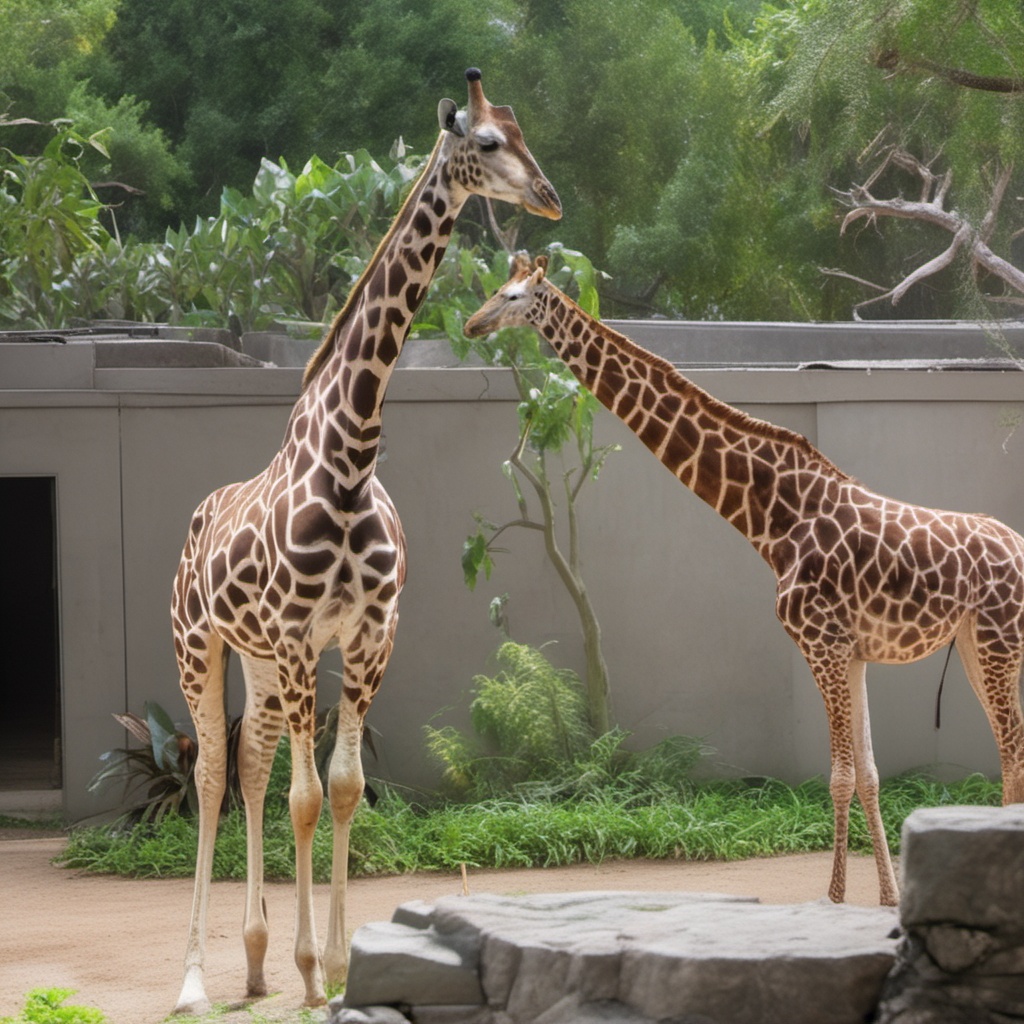}
    
  \end{subfigure}%
  \hfill 
  \begin{subfigure}{.48\linewidth}
    \centering
    \includegraphics[width=\linewidth]{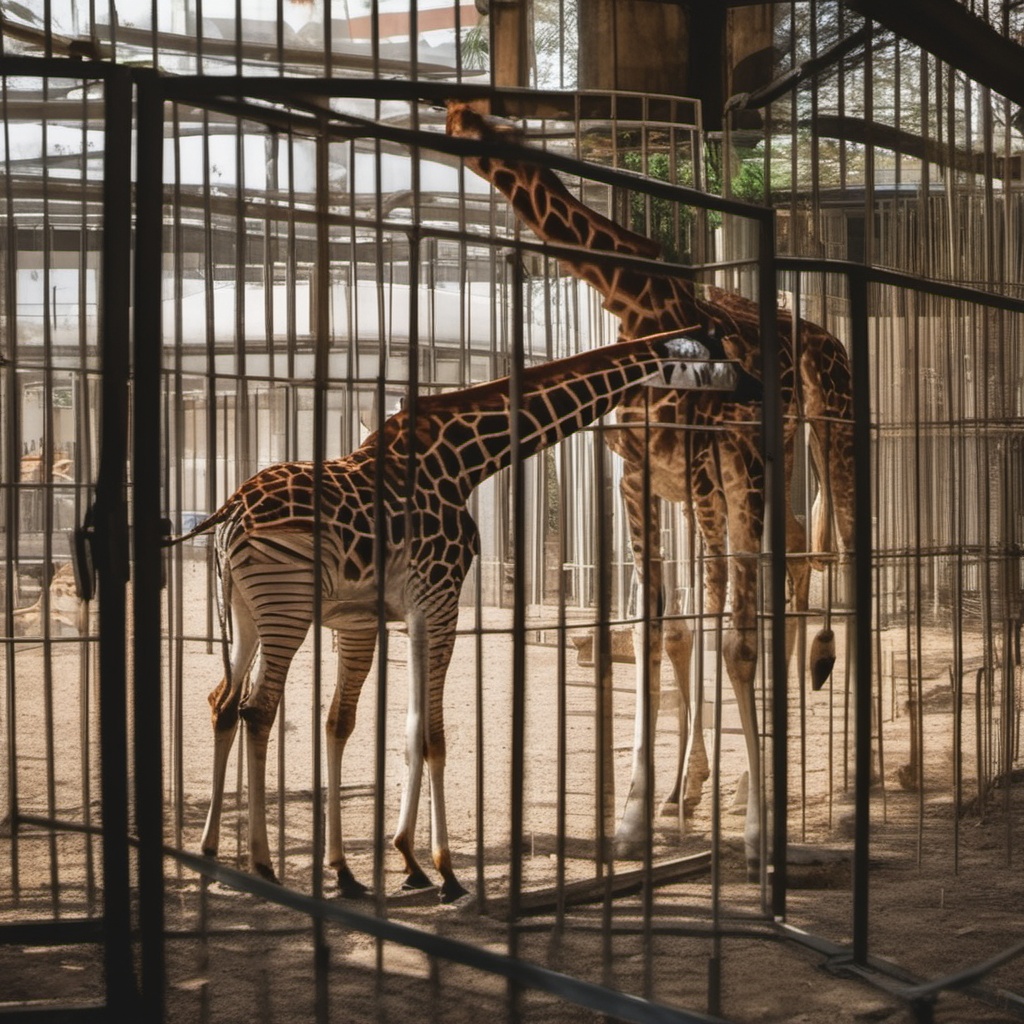}
    
  \end{subfigure}
    (g)
\end{minipage}%
\hfill
\begin{minipage}{.22\textwidth}
  \centering
  \begin{subfigure}{.48\linewidth}
    \centering
    \includegraphics[width=\linewidth]{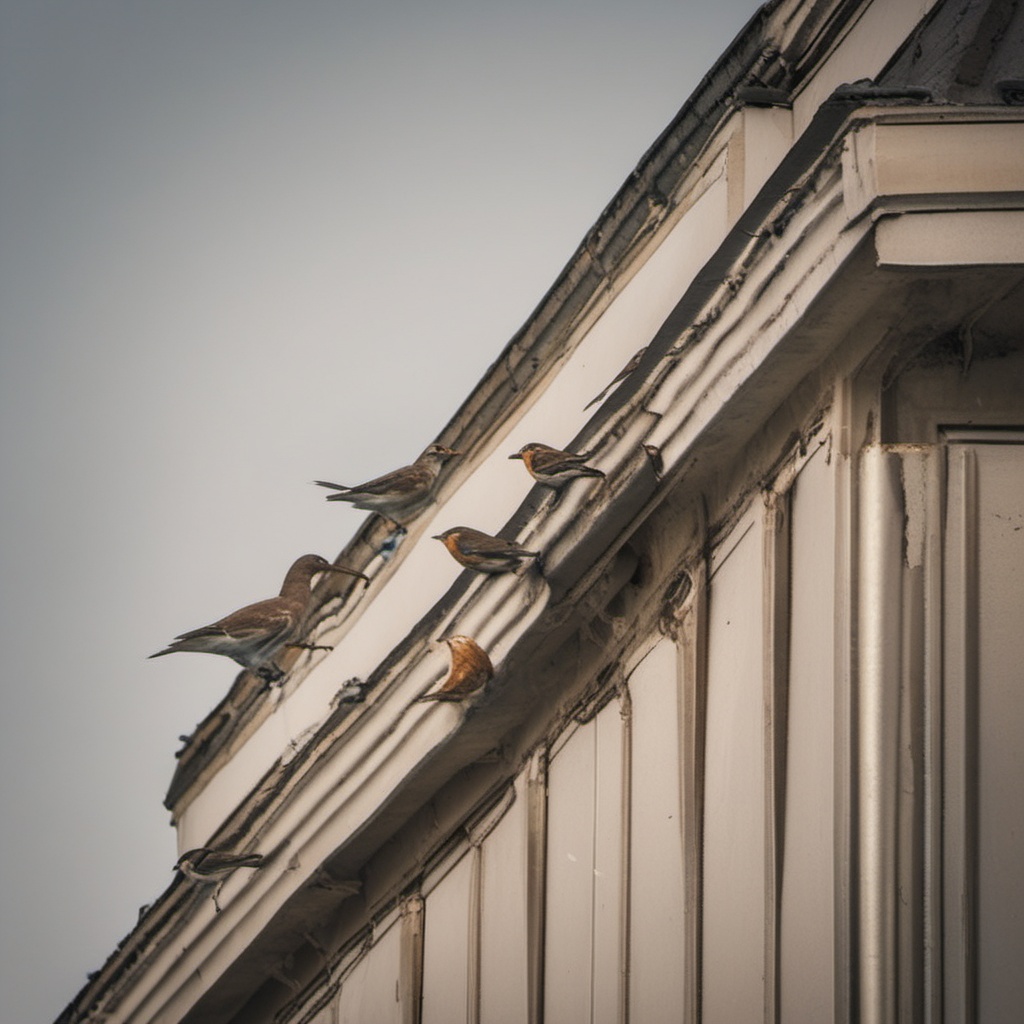}
    
  \end{subfigure}%
  \hfill 
  \begin{subfigure}{.48\linewidth}
    \centering
    \includegraphics[width=\linewidth]{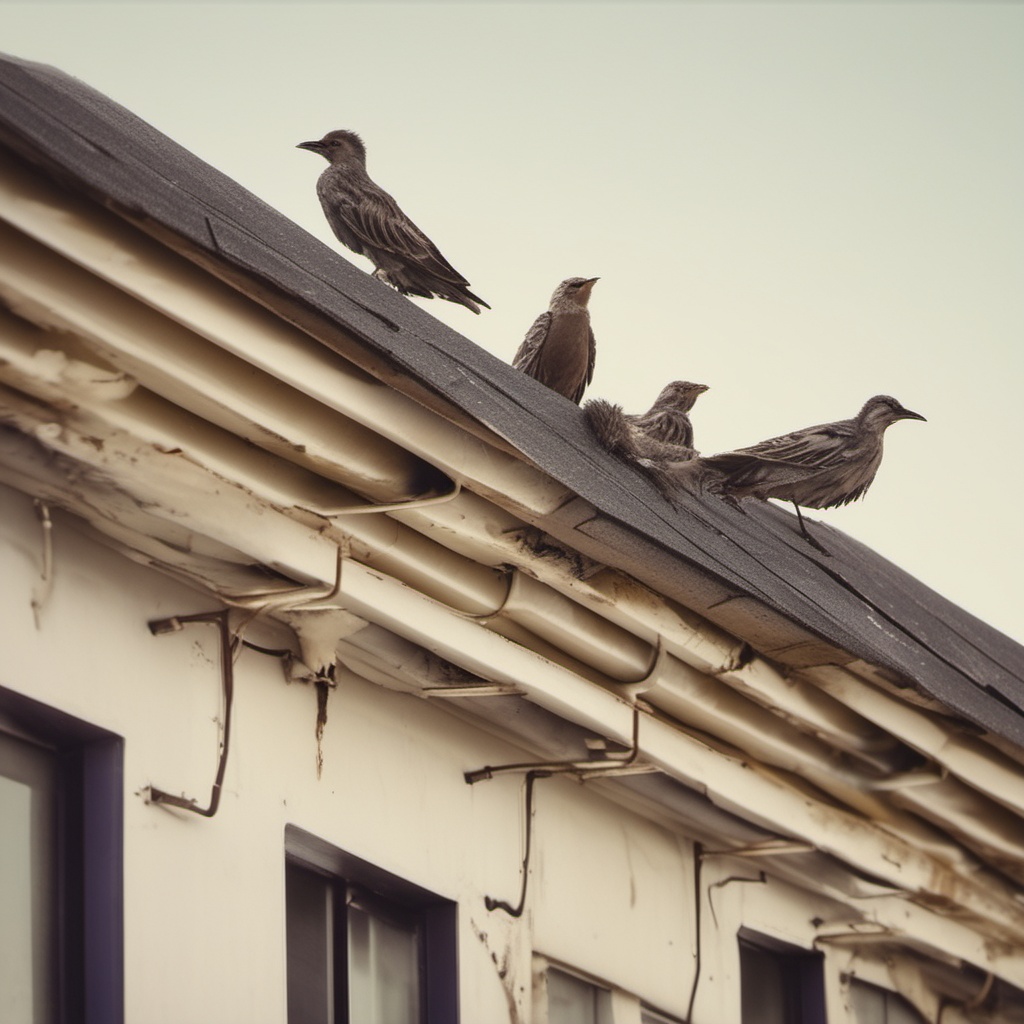}
    
  \end{subfigure}\\
  \begin{subfigure}{.48\linewidth}
    \centering
    \includegraphics[width=\linewidth]{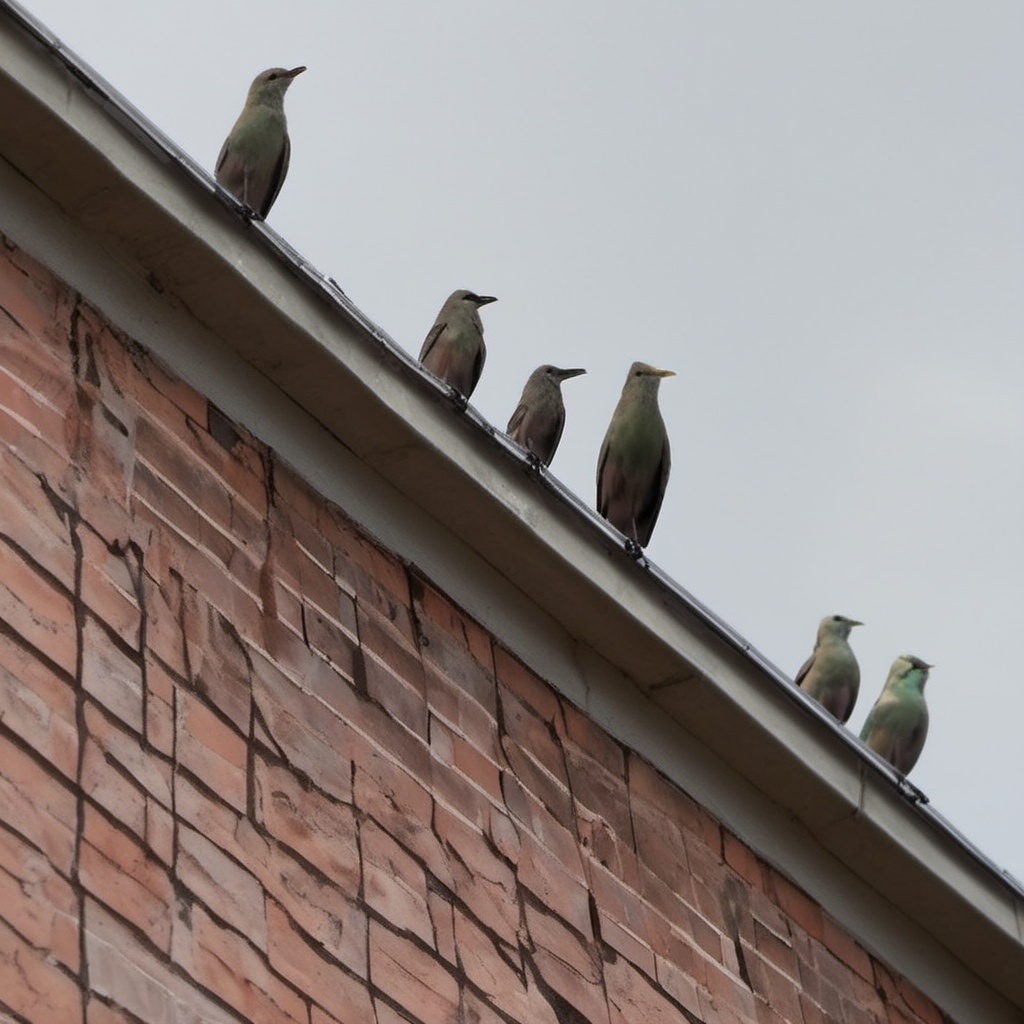}
    
  \end{subfigure}%
  \hfill 
  \begin{subfigure}{.48\linewidth}
    \centering
    \includegraphics[width=\linewidth]{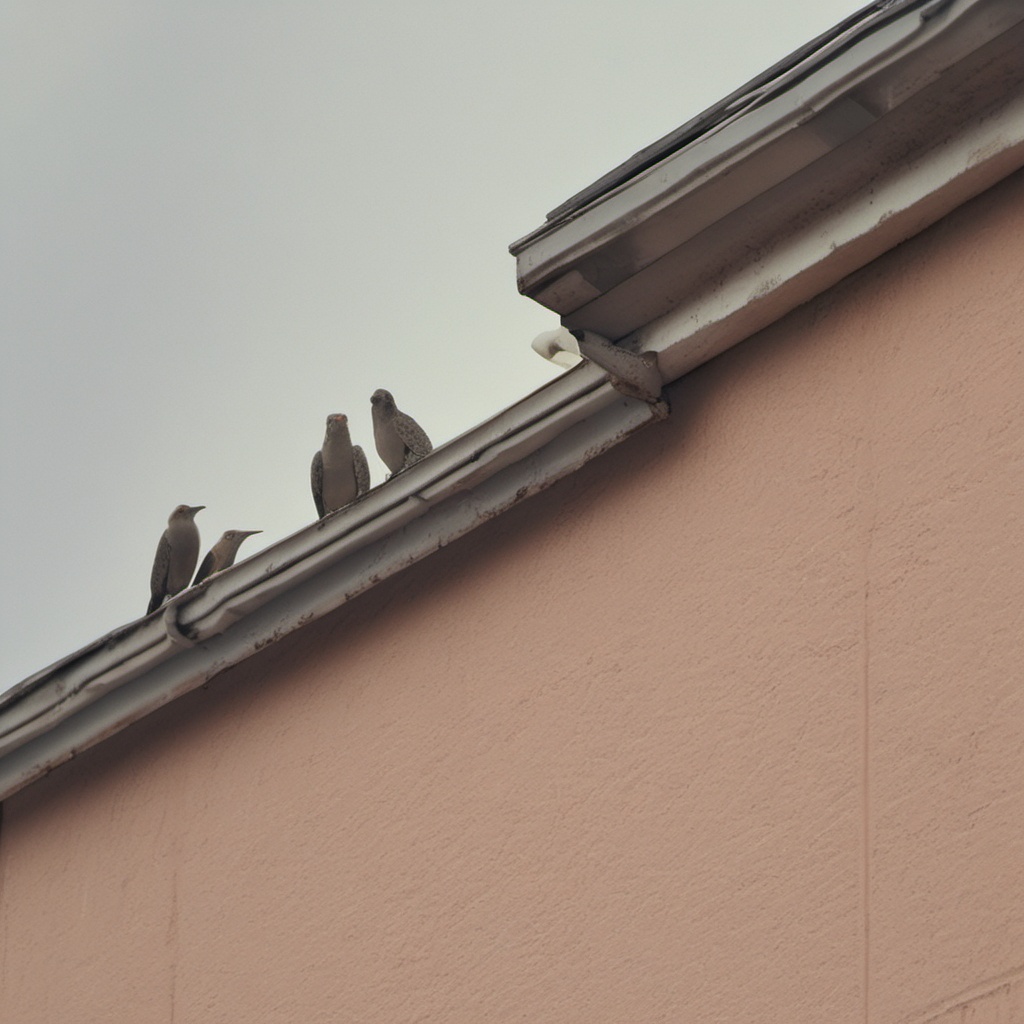}
    
  \end{subfigure}
    (h)
\end{minipage}%
\vspace{0.1em} 

\rotatebox[origin=c]{90}{Ours}\quad 
\begin{minipage}{.22\textwidth}
  \centering
  \begin{subfigure}{.48\linewidth}
    \centering
    \includegraphics[width=\linewidth]{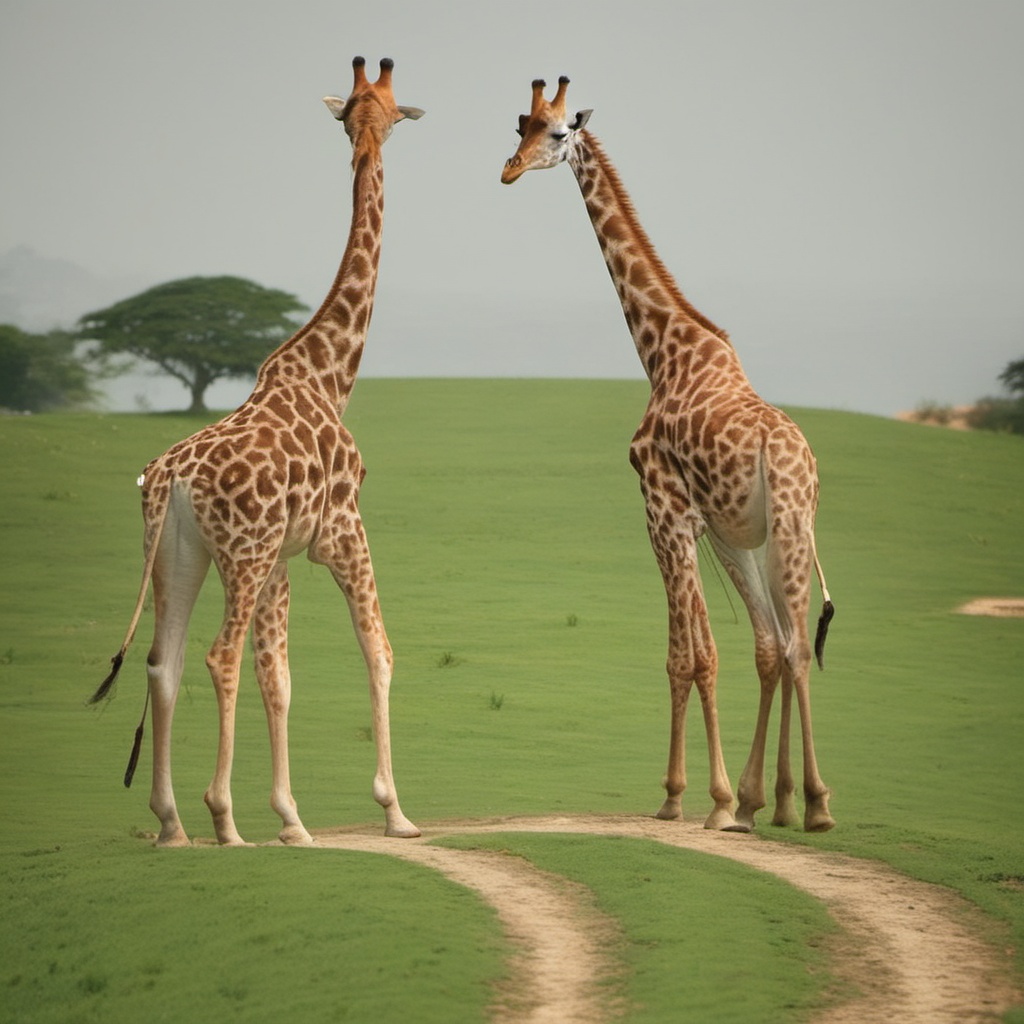}
    
  \end{subfigure}%
  \hfill 
  \begin{subfigure}{.48\linewidth}
    \centering
    \includegraphics[width=\linewidth]{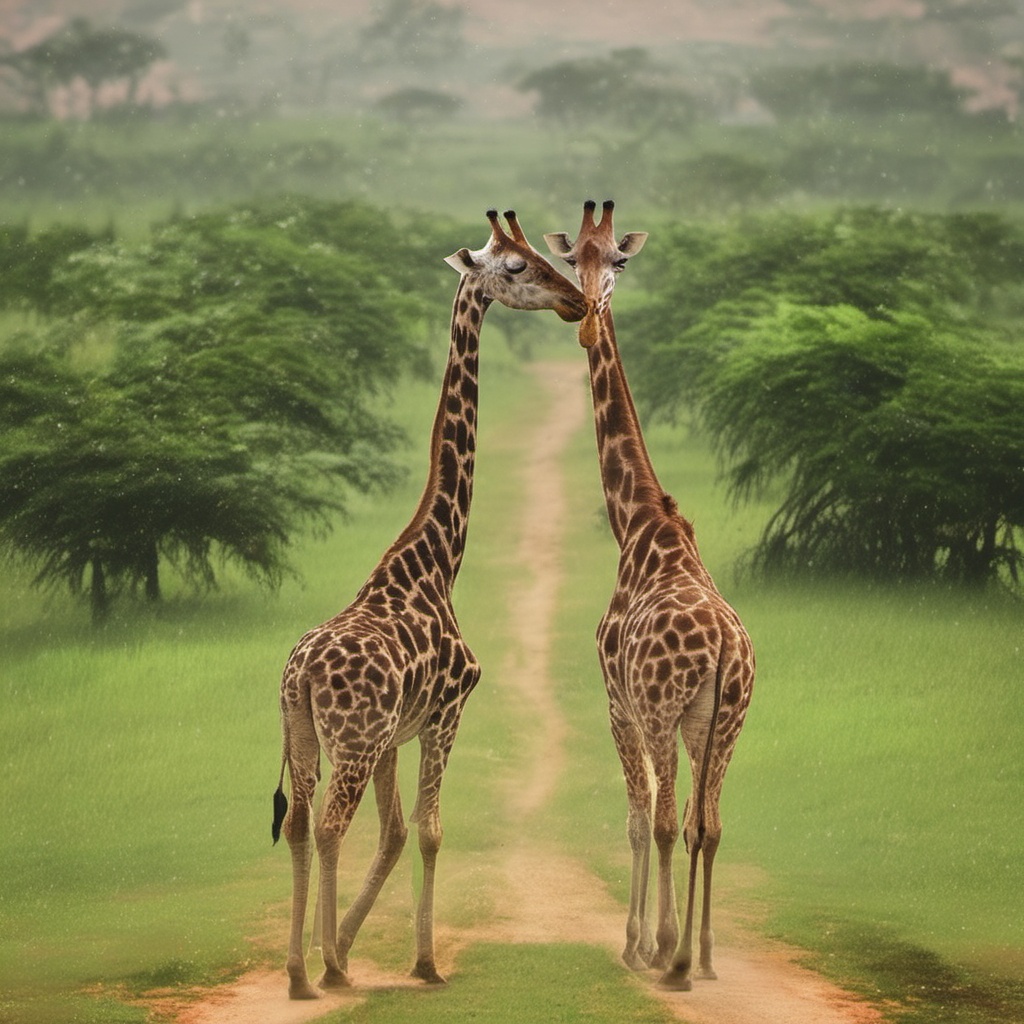}
    
  \end{subfigure}\\
  \begin{subfigure}{.48\linewidth}
    \centering
    \includegraphics[width=\linewidth]{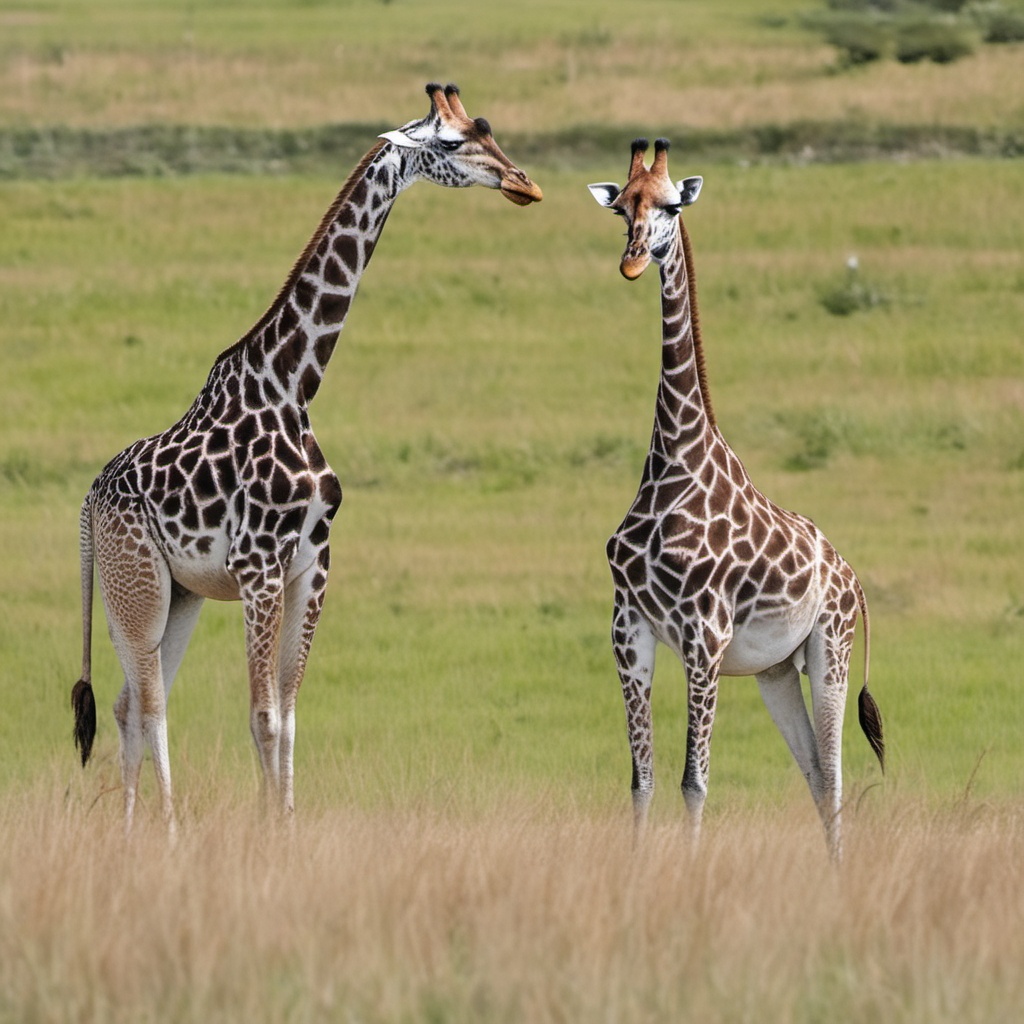}
    
  \end{subfigure}%
  \hfill 
  \begin{subfigure}{.48\linewidth}
    \centering
    \includegraphics[width=\linewidth]{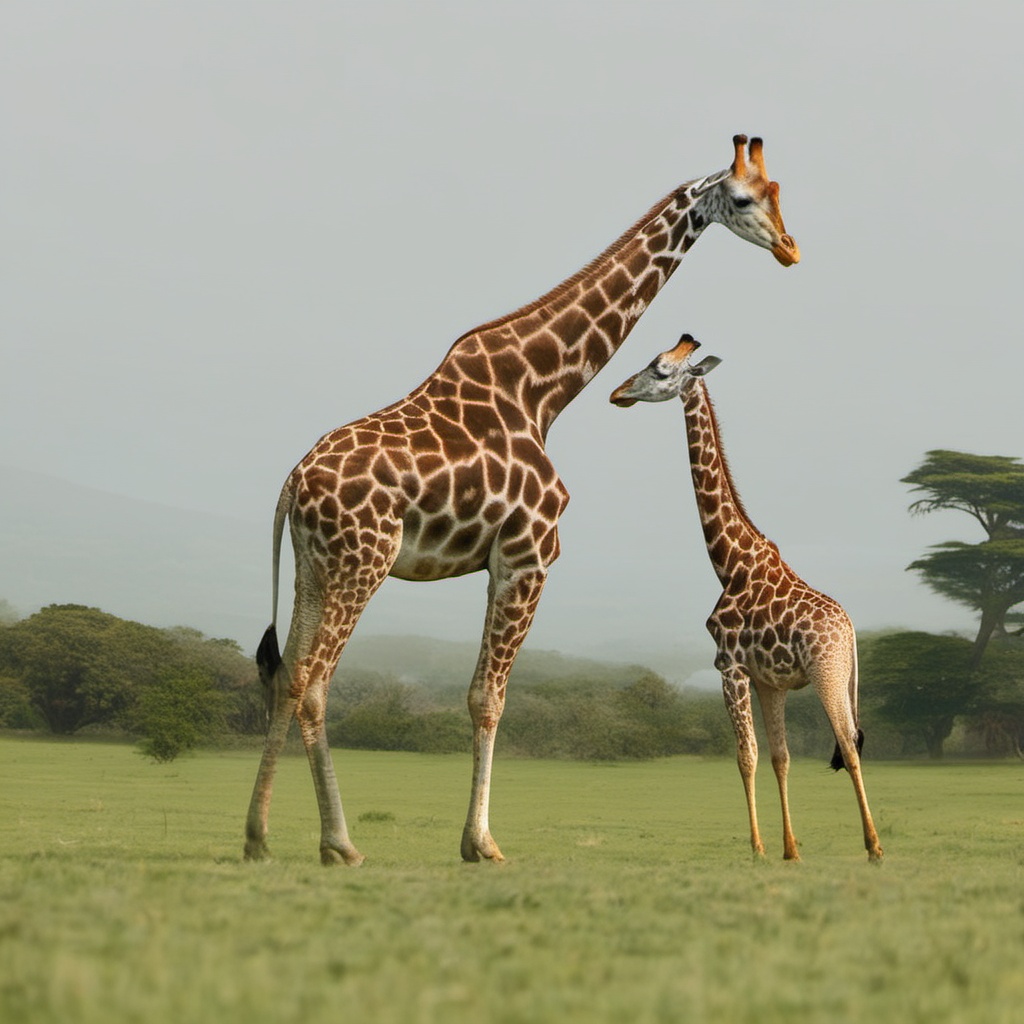}
    
  \end{subfigure}
    (i)
\end{minipage}%
\hfill 
\begin{minipage}{.22\textwidth}
  \centering
  \begin{subfigure}{.48\linewidth}
    \centering
    \includegraphics[width=\linewidth]{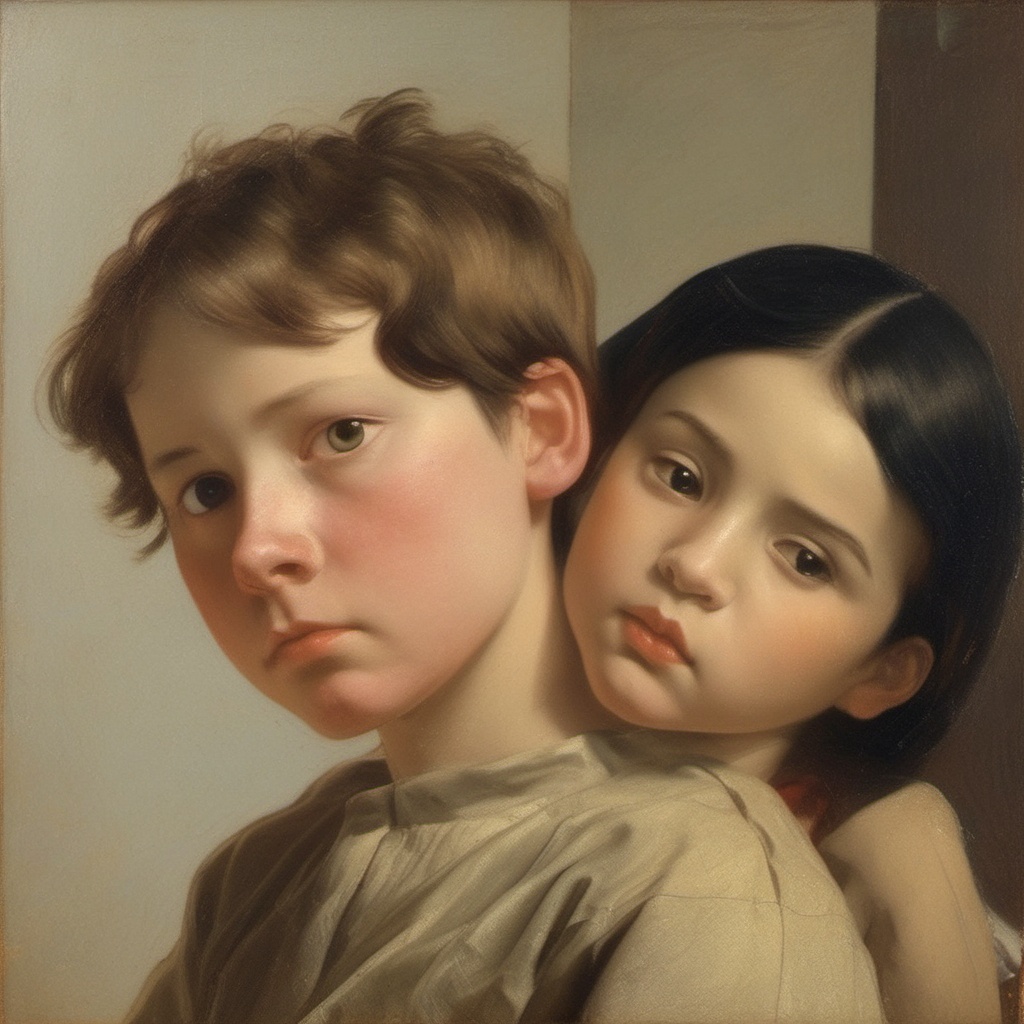}
    
  \end{subfigure}%
  \hfill 
  \begin{subfigure}{.48\linewidth}
    \centering
    \includegraphics[width=\linewidth]{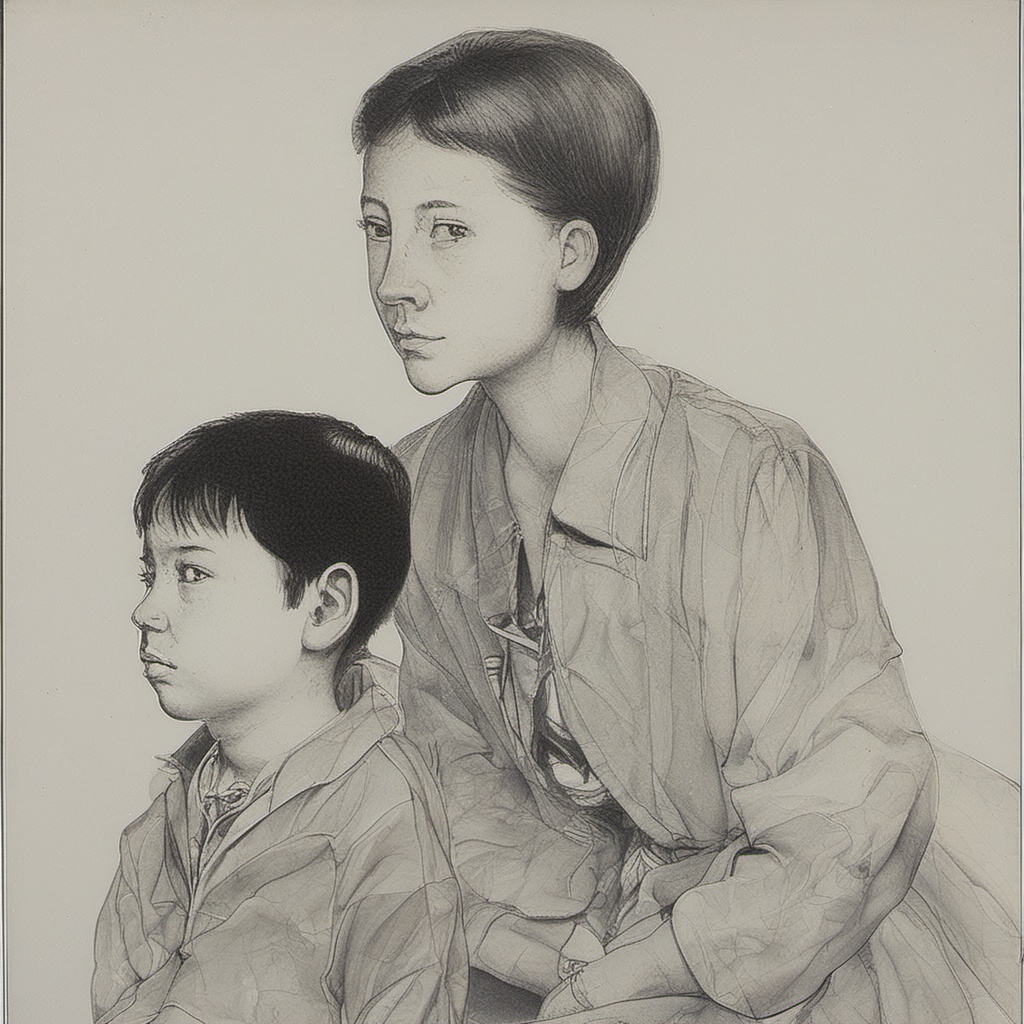}
    
  \end{subfigure}\\
  \begin{subfigure}{.48\linewidth}
    \centering
    \includegraphics[width=\linewidth]{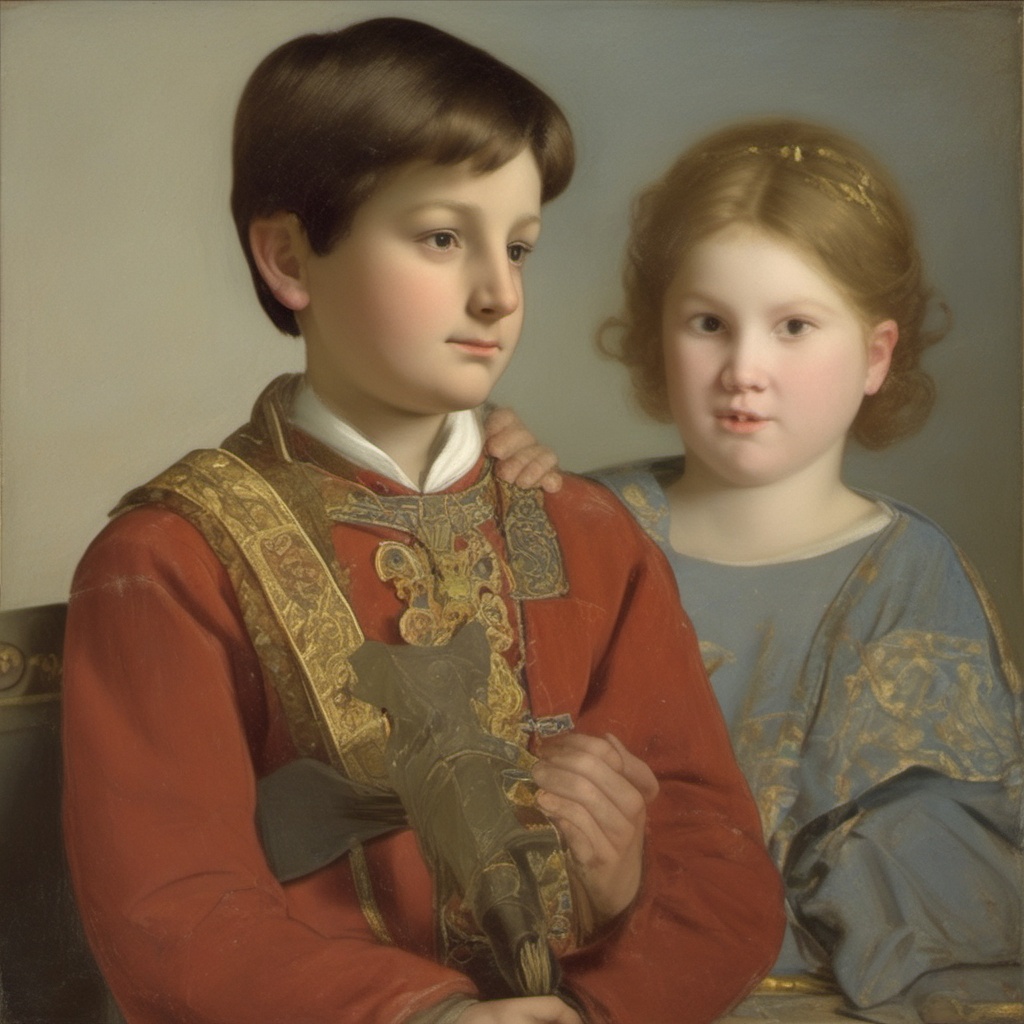}
    
  \end{subfigure}%
  \hfill 
  \begin{subfigure}{.48\linewidth}
    \centering
    \includegraphics[width=\linewidth]{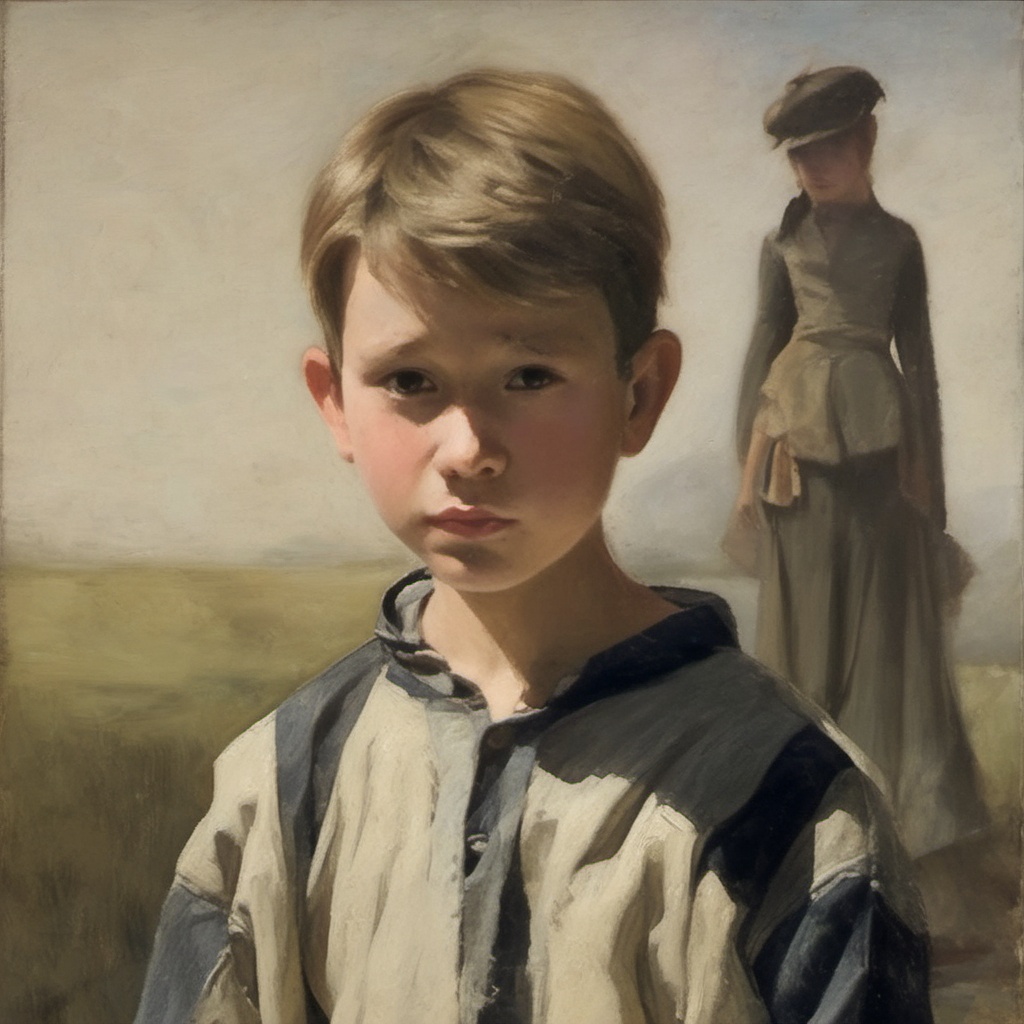}
    
  \end{subfigure}
    (j)
\end{minipage}%
\hfill
\begin{minipage}{.22\textwidth}
  \centering
  \begin{subfigure}{.48\linewidth}
    \centering
    \includegraphics[width=\linewidth]{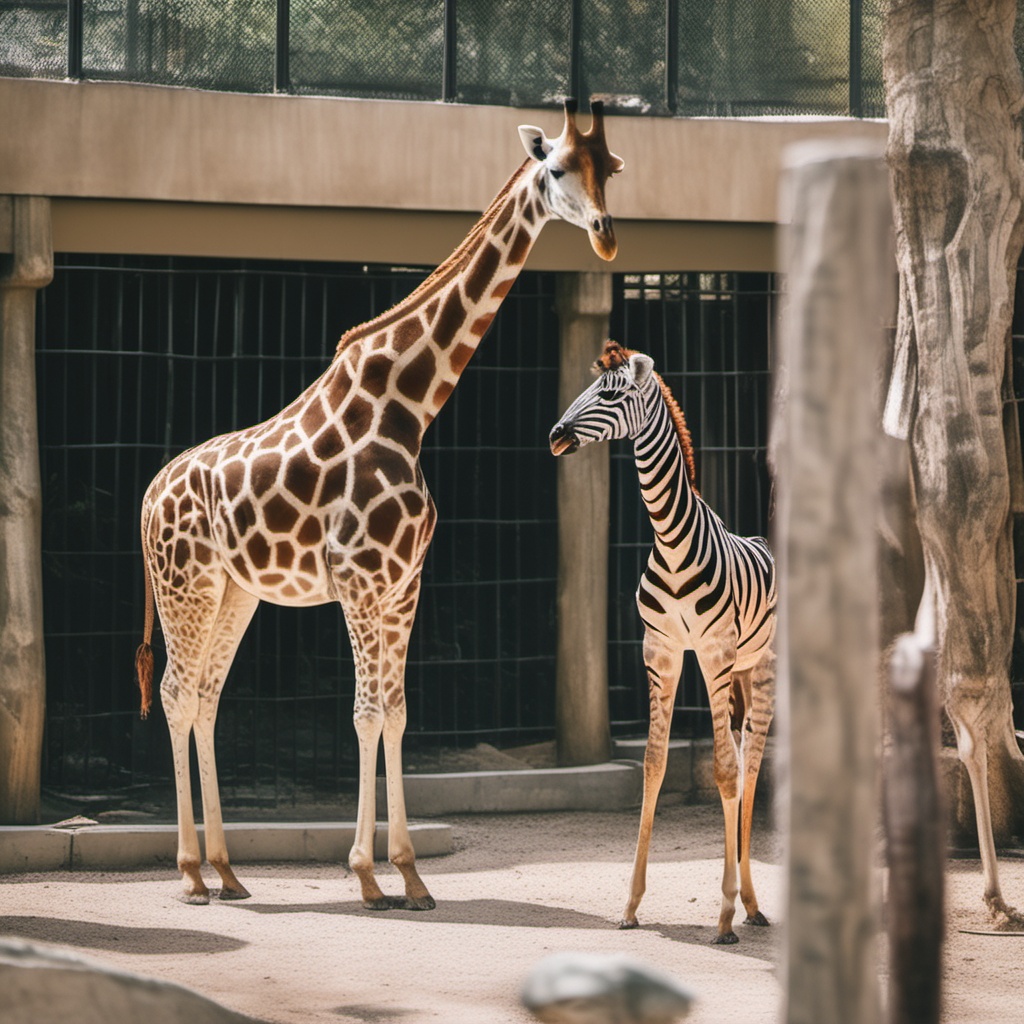}
    
  \end{subfigure}%
  \hfill 
  \begin{subfigure}{.48\linewidth}
    \centering
    \includegraphics[width=\linewidth]{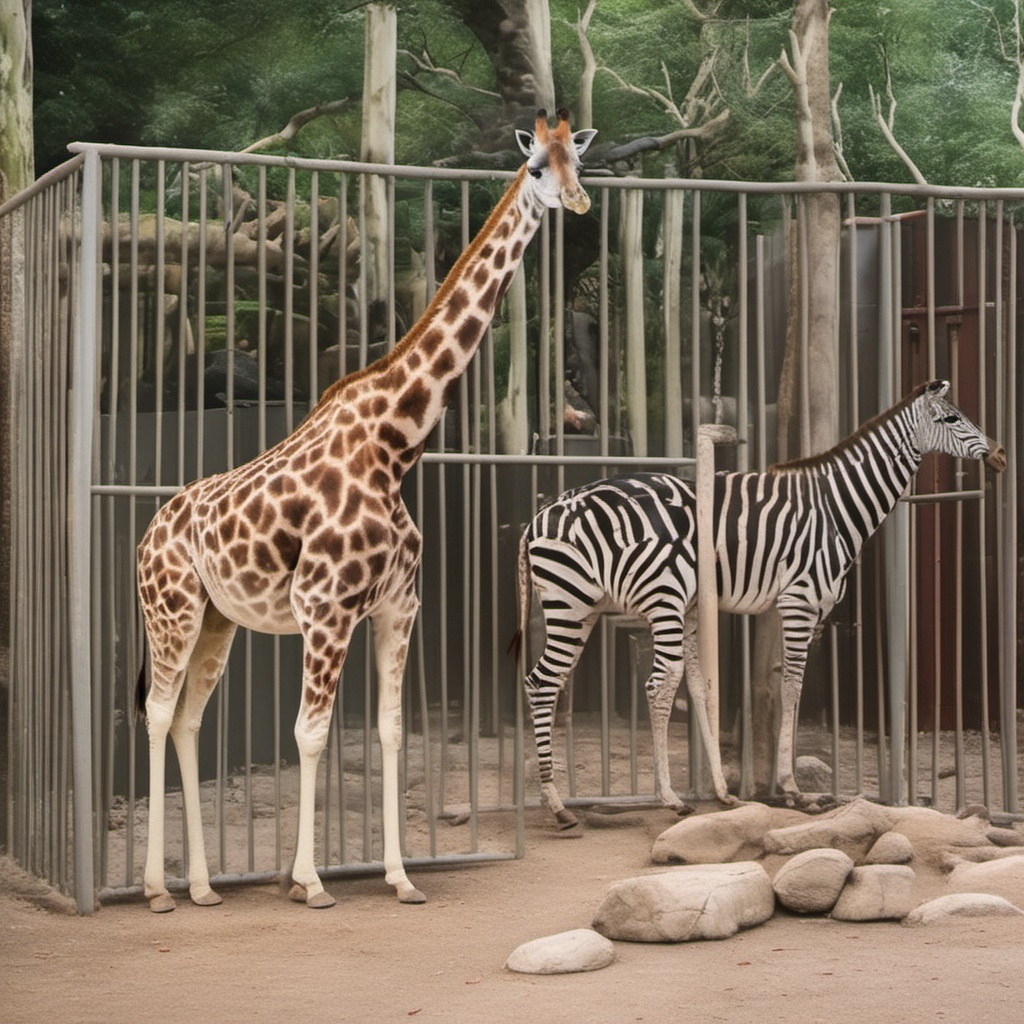}
    
  \end{subfigure}\\
  \begin{subfigure}{.48\linewidth}
    \centering
    \includegraphics[width=\linewidth]{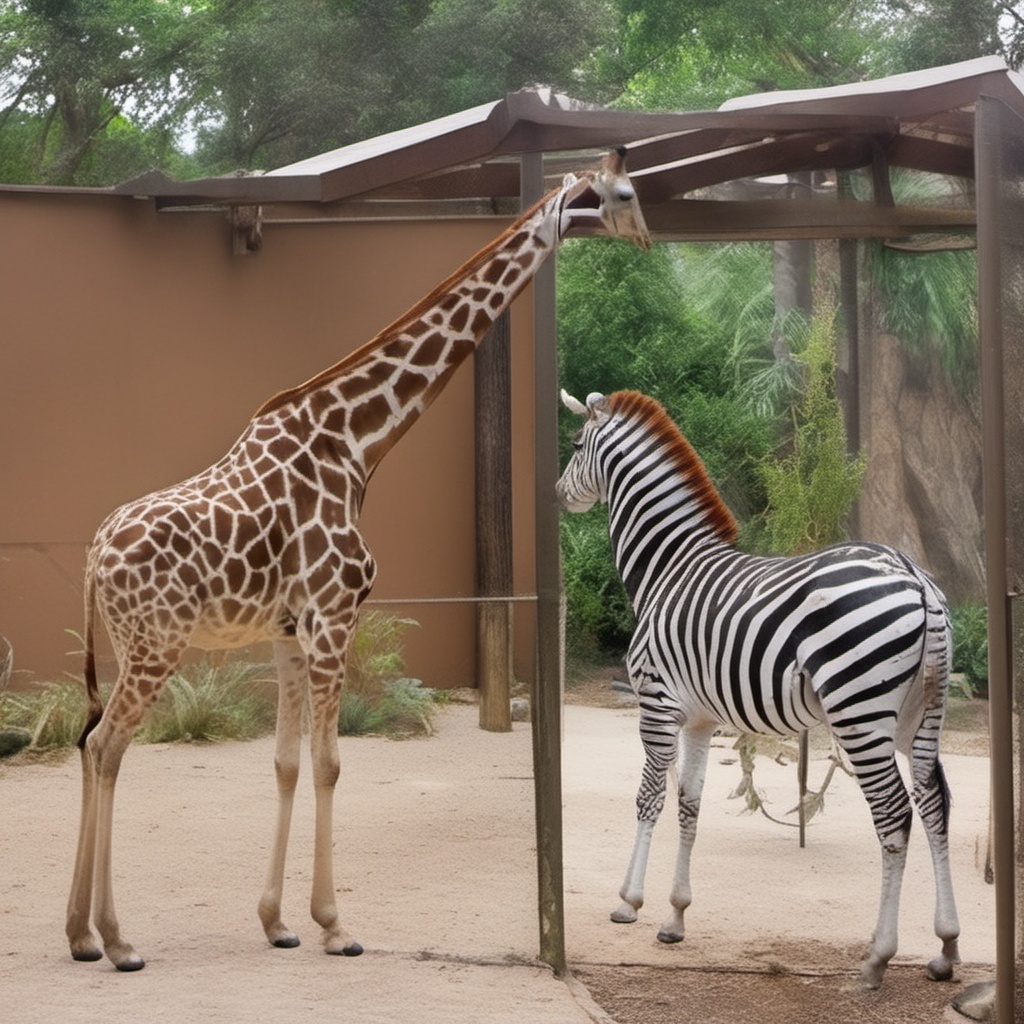}
    
  \end{subfigure}%
  \hfill 
  \begin{subfigure}{.48\linewidth}
    \centering
    \includegraphics[width=\linewidth]{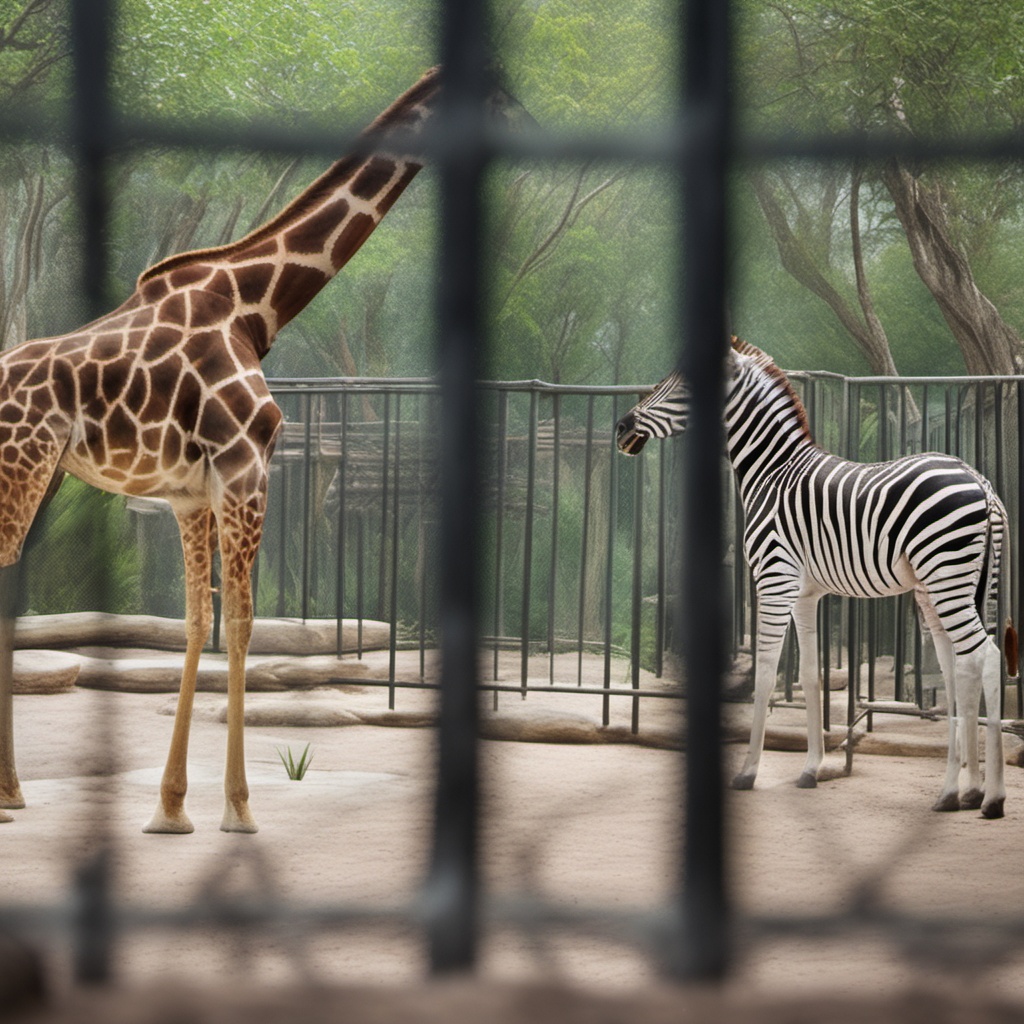}
    
  \end{subfigure}
  (k)
\end{minipage}%
\hfill
\begin{minipage}{.22\textwidth}
  \centering
  \begin{subfigure}{.48\linewidth}
    \centering
    \includegraphics[width=\linewidth]{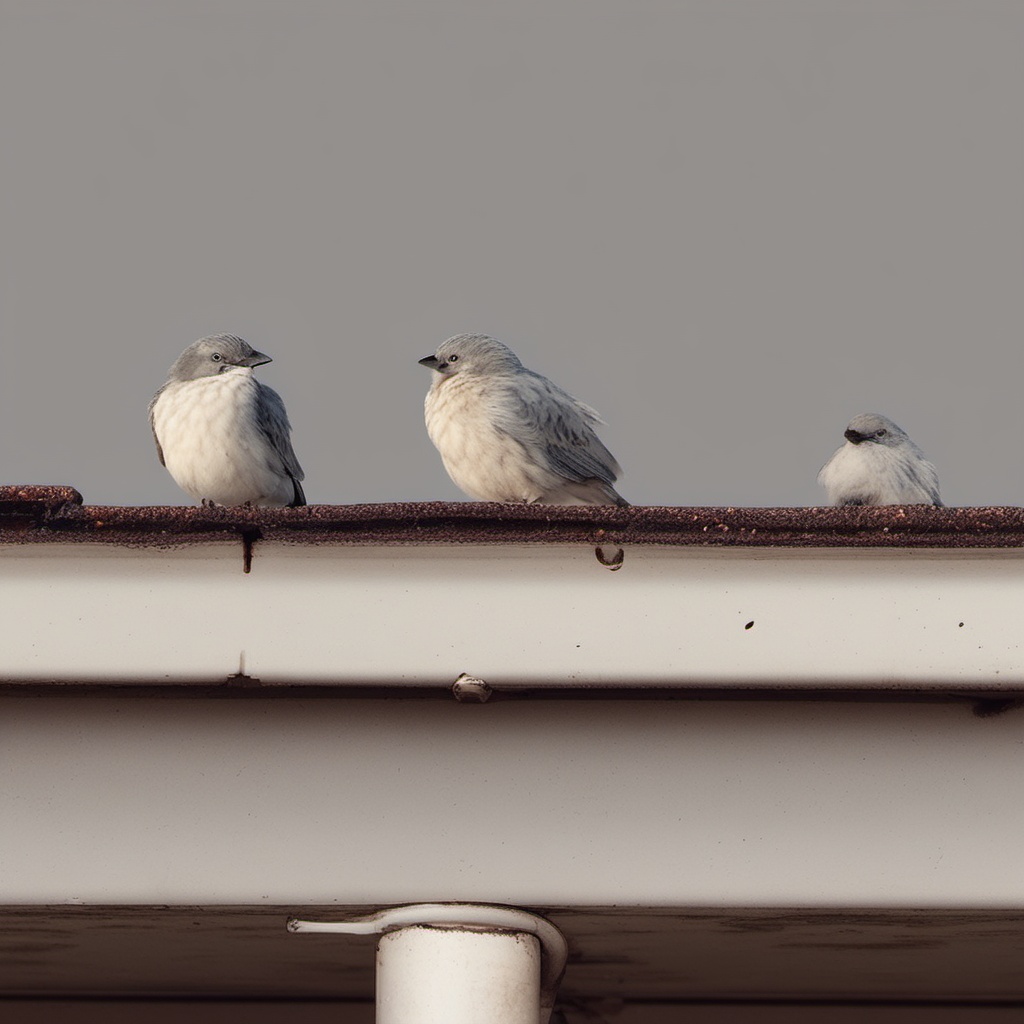}
    
  \end{subfigure}%
  \hfill 
  \begin{subfigure}{.48\linewidth}
    \centering
    \includegraphics[width=\linewidth]{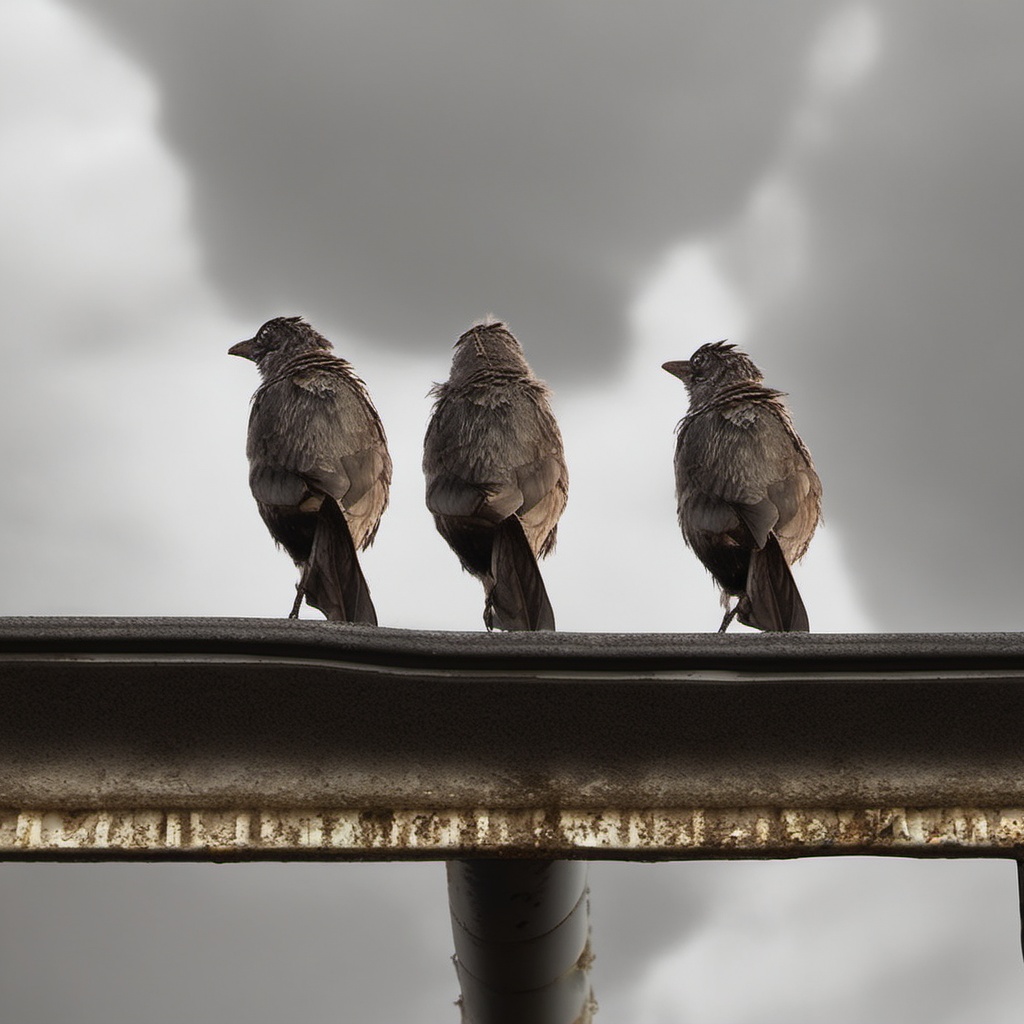}
    
  \end{subfigure}\\
  \begin{subfigure}{.48\linewidth}
    \centering
    \includegraphics[width=\linewidth]{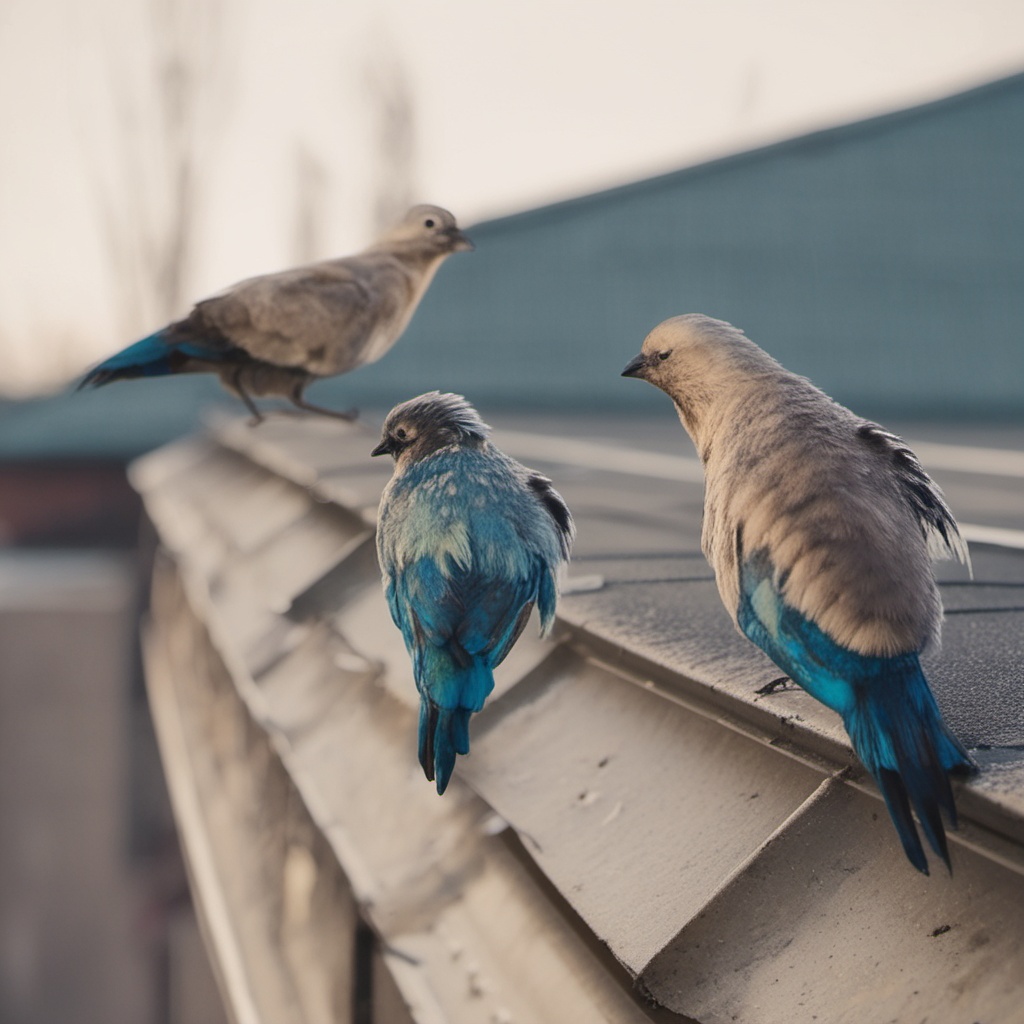}
    
  \end{subfigure}%
  \hfill 
  \begin{subfigure}{.48\linewidth}
    \centering
    \includegraphics[width=\linewidth]{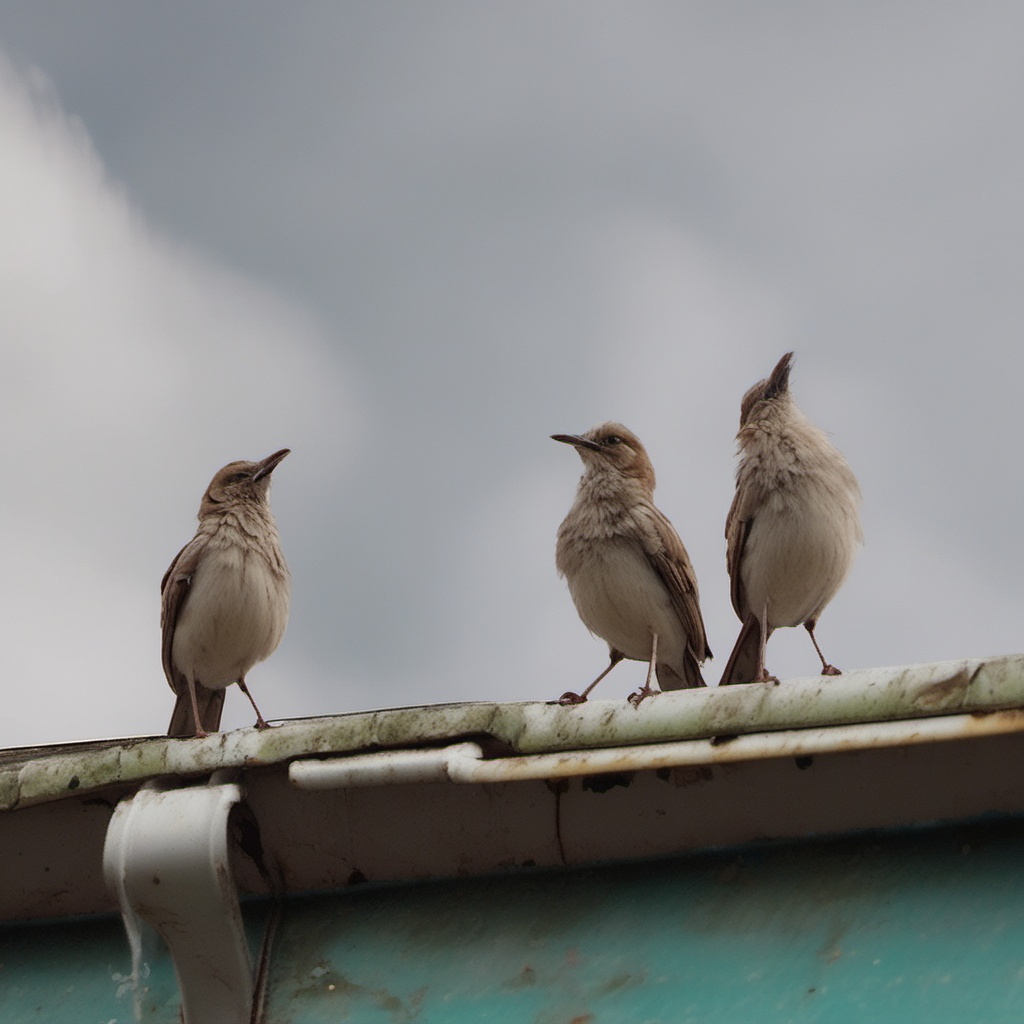}
    
  \end{subfigure}
  (l)
\end{minipage}%
\caption{\textbf{The qualitative results given the prompt.} In our qualitative experiments, we conducted extensive tests on Stable Diffusion XL, Structured Diffusion, and our method. The results indicate that our approach generates images that are closer to the prompt, and more accurately align in terms of quantity, attributes, and object alignment, as per expectations.}
\label{qualitative}
\end{figure*}

\subsection{Qualitative Analysis}
\label{41}
In Figure \ref{qualitative}, we showcase the qualitative results obtained from Stable Diffusion, Structured Diffusion, and our method. It is observed that existing models still struggle with multiple alignment issues. 
Focusing on the issue of numerical alignment, as demonstrated in the first and fourth columns in Figure \ref{qualitative}, we can observe that existing models, including Stable Diffusion and Structured Diffusion, often fail to accurately replicate the specified quantities in the prompts. In (a)(d)(e)(h), there is a noticeable excess in the number of giraffes and birds generated compared to the prompt's specification, while it can be observed from (i)(l) that our method can alleviate the occurrence of such numerical errors. 

The case when prompts involve multiple entities is shown in the second and third columns. For Stable Diffusion XL, there is a noticeable numerical misalignment, as seen in the first two images of (b). Additionally, attribute leakage is also evident in the last two images of (b),  where, despite the correct quantity, the generated images depict only boys. A similar issue is observed in the prompts involving giraffes and zebras in (c); the images correctly portray the number and setting, yet the giraffe's characteristics dominate, leading to a zebra that inappropriately exhibits giraffe-like features.
In Structured Diffusion, such issues have been somewhat ameliorated. In (f), the model generates the anticipated scenes, yet it continues to encounter problems of attribute leakage and entity disappearance. Meanwhile, in (g), there has been notable progress in addressing attribute leakage problem, but the misalignment of attributes is still an evident issue. In the first image in (g), the attributes of zebra and giraffe have been successfully separated, but they are misplaced, resulting in a distorted image. Results in (j) and (k) indicate that our method effectively distinguishes and generates distinct entities while simultaneously diminishing the frequency of attribute leakage.
\subsection{Quantitative Analysis}
\label{42}
\subsubsection{Benchmark Evaluation}
\begin{figure}[ht] 
  \centering
  \includegraphics[width=0.49\textwidth]{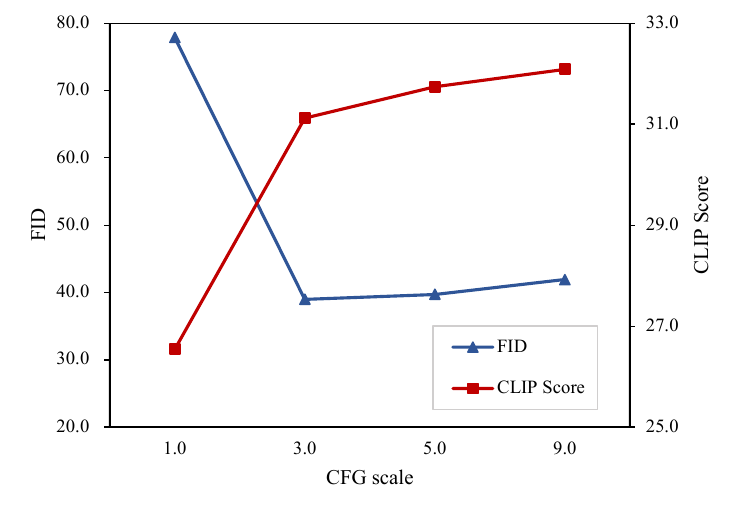} 
  \caption{\textbf{FID and CLIP Score of our method under different CFG scales.} CFG refers to the parameter in Stable Diffusion that controls the degree of relevance between the final generated text and the input prompt. We tested the model's performance under different CFG settings.}
  \label{diff_cfg}
\end{figure}

We first test the performance of our model on the COCO validation set across various CFG scales, with the result shown in Figure \ref{diff_cfg}. The CFG scale is a hyperparameter to govern the extent of text prompts and the extent of text prompts' impact on image generation. We find that our model attains a balanced performance between FID and CLIP scores when CFG is set to 3 and 5. Consequently, for subsequent experiments, we standardize the CFG at 5 to facilitate comparative analyses with other models.

Then we conduct comparison experiments with different methods, including our backbone Stable Diffusion XL~\cite{podell2023sdxl} and Structured Diffusion~\cite{feng2022training}.
The experimental results are presented in Table \ref{comparison}. 
\begin{table}[]
\centering
\caption{\textbf{The quantitative comparison of FID, CLIP and ImageReward Score with other methods.} We also conducted the CLIP score evaluation for ground-truth images in the dataset. Based on these scores, we discussed the credibility of the two evaluation criteria below.} 
\begin{tabular}{@{}cccc@{}}
\toprule
Model& FID $\downarrow$ & CLIP Score $\uparrow$ & ImageReward $\uparrow$ \\ 
\midrule
Stable Diffusion XL & 40.37 & 31.70 & 6.66 \\
Structured Diffusion&41.18 &31.50 & 5.91\\
Ground-Truth& - &30.32 & - \\
\textbf{Ours} & \textbf{39.71} &\textbf{31.74} & \textbf{6.68} \\
\bottomrule
\end{tabular}
\label{comparison} 
\end{table}
It can be observed that our model can generate images that more closely resemble ground-truth images and exhibit higher alignment with the input prompts when compared to the other methods. It is noteworthy that in the third line of the experiment, an assessment was conducted on the CLIP Score of the ground-truth images in dataset with their corresponding caption annotations. However, the results are suboptimal. We believe there could be two reasons: (1) the CLIP Score is not completely reliable and should not be considered an absolute measure of performance;(2)The annotations within the COCO2014 dataset do not sufficiently or accurately reflect the visual semantic information. Hence, text embedding map to the latent space is less similar to the ground-truth visual embedding and consequently attains lower CLIP scores. Additionally, we believe that the qualitative results of FID also fail to reflect the actual generative capability of these models, as FID compares the similarity between the generated images and the ground-truth images. It overlooks that a prompt can be represented in various visual forms (i.e., various scenarios can be interpreted by the same description), not limited to a single ground-truth picture. These insights are instrumental in informing the direction of our future research endeavors.

\subsubsection{Semi-human Evaluation for Alignment Performance}
In addition to FID, CLIP and ImageReward Score metrics, we conduct human evaluation with the aid of GPT-4~\cite{openai2023gpt4}. In this experiment, we randomly selected 50 prompts and generated images using baselines and our model. For the ground-truth prompt, we feed it into GPT-4 and let it generate 135 questions from different perspectives such as alignment, activity, and detail.

 The results are shown in Table \ref{clean_gpt}. GPT-4 primarily formulated questions from the following five aspects:
\begin{itemize}
    \item \textbf{Color Alignment}: These questions primarily examine whether the colors specified for different objects in the prompt have been accurately generated in the image. 
    \item \textbf{Content Alignment}: These questions mainly concentrate on the scenes, objects, and character actions in the image, scrutinizing whether the content generated is consistent with prompt.
    \item \textbf{Numerical Alignment}: These questions investigate whether the image accurately generates the specified quantities of different objects in the prompt. 
    \item \textbf{Surface/Texture Alignment}: This mainly examines whether the objects generated possess textures that are consistent with those described in the prompt. 
    \item \textbf{Time Alignment}: This refers to whether the environmental time presented in the image matches the time described in the prompt.
    \item \textbf{Location}: This evaluates whether the environmental settings and spatial relationships generated are consistent with the prompt.
\end{itemize}

We prompt GPT-4 Vision to describe the generated images and then manually check whether these questions were correctly addressed or not. Then we manually inspected the questions and corrected them.

\begin{table}[]
    \centering
    \caption{\textbf{Human evaluation accuracy on different tasks (\%).} We manually revised the comparison results based on GPT-4's answers. Our model has achieved outstanding results in all alignment tasks.}
    \begin{tabular}{cccc}
\toprule
Question Type             & Ours          & Stable Diffusion XL & Structured Diffusion \\ \midrule
Color Alignment           & \textbf{85.71}         & 71.43               & 76.19                     \\
Context Alignment         & \textbf{90.00}         & 70.00               & 70.00                     \\
Numerical Alignment       & \textbf{57.14}         & 50.00               & 46.43                     \\
Surface/Texture Alignment & \textbf{100.00}        & 80.00               & \textbf{100.00}                    \\
Time Alignment & \textbf{100.00}        & \textbf{100.00}               & \textbf{100.00}                    \\
Location                  & \textbf{80.00}         & 60.00               & 46.67                     \\ \bottomrule
\end{tabular}
    \label{clean_gpt}
\end{table}
It can be observed in Table \ref{clean_gpt} that, \textbf{our model achieved the best performance in all alignment tasks} compared with Stable Diffusion XL and Structured Diffusion. \textbf{In color, context, numerical alignment, and location tasks, our performance is significantly superior to Structured Diffusion}. Such results are consistent with the qualitative and quantitative experimental outcomes, demonstrating that our proposed paradigm can indeed effectively enhance stable diffusion's attribute alignment capability.
\subsection{Ablation Experiment}
\label{43}
We also conduct ablation studies to assess the impact of three key components on our method: the Self-Attention control strategy, the object-focused masking mechanism, and the dynamic reweighting strategy. The results are shown in Table \ref{tab:ablation}. 
\begin{table}
\caption{\textbf{The ablation result of FID and CLIP Score.} Both the self-attention control strategy and the object-focused mask have certain improvements over the baseline. By integrating these three components, our model achieve more robust performance on FID and CIP Score metrics.}
\centering
\begin{tabular}{@{}ccc@{}}
\toprule
Model& FID $\downarrow$ & CLIP Score $\uparrow$ \\ 
\midrule
Stable Diffusion XL & 40.37 & 31.70 \\
Only Self-Attention control&40.34 &31.83\\
Only Object-focused Mask& 39.55 &31.71 \\
Only Dynamic Reweighting& 49.7 &27.18 \\
Ours & 39.71 &31.74 \\
\bottomrule
\end{tabular}
\label{tab:ablation} 
\end{table}
It can be found that both the self-attention control strategy and the object-focused mask have certain improvements over the baseline. However, the dynamic reweighting strategy on its own does not perform well in the ablation studies. The reason lies in the unmasked noises. In this ablation setting, the original attention distribution is emphasized without masking, which also emphasizes the redundant distribution, thus the generated effect is not as expected. In fact, the mask mechanism and dynamic weighting mechanism complement each other to work well. After using the mask to remove the redundant subject information in the attention distribution, further phase-wise emphasis can enable the model to focus better on specific semantic components. For more detailed ablation experiments for different components of our model, please refer to our supplementary materials.

\section{Discussion}

\begin{figure}
\centering
\includegraphics[width=0.6\linewidth]{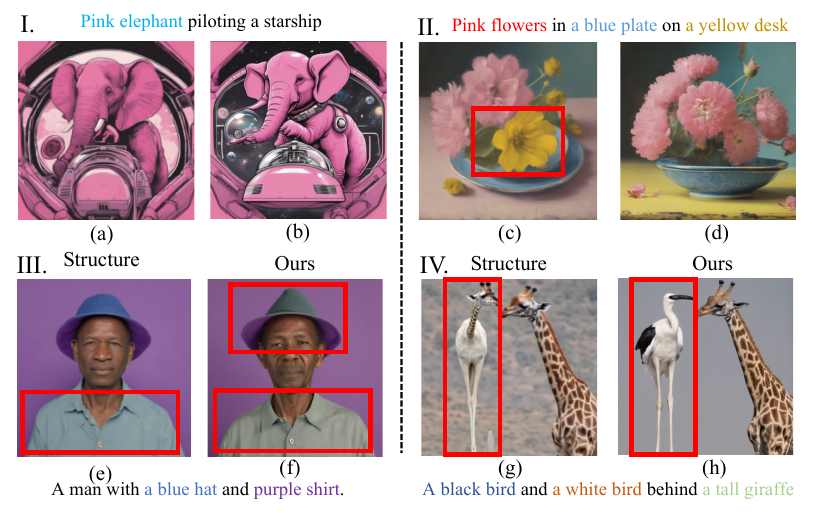}
\caption{\textbf{Error analysis}: We conduct the error analysis experiments for Structured Diffusion and ours, the result could be divided into four groups: I)Single Object-Attribute pair, II) Multiple Object-Attribute pairs, III) Nested Objects and attributes, IV) Same object with different attributes.}
\label{error_analysis}
\end{figure}
We provide an error analysis as shown in Figure \ref{error_analysis}. We divide the tasks into four categories:

\textbf{Single Object-Attribute pair.} In (a) and (b), \textbf{both} our method and Structured Diffusion \textbf{perform well} in scenarios of only one object-attribute pair. 

\textbf{Multiple Object-Attribute pairs.} In (c) and (d), where the prompt contains multiple entities and attributes, \textbf{our model performs better}. This is because Structured Diffusion can not guarantee the semantic information is only superimposed onto the corresponding regions of the image. In (c), the semantic of a yellow table might leak into the region of the flower above it, which misguides the model to create a yellow flower, resulting in attribute misalignment. 

\textbf{Nested Objects and attributes.} This refers to a hierarchical relationship in the prompt, where a primary entity encompasses several sub-entities with corresponding attributes. In (e) and (f), both Structured Diffusion and our method failed to achieve the expected alignment. This is due to the scenario where one entity corresponds to multiple sub-entities. For instance, a man's attributes include a hat and a T-shirt, each with its own attributes. \textbf{Both two model exhibits degradation} in cases where attributes and entities are nested, which might be related to our shared backbone, Stable Diffusion itself. 

\textbf{Same object with different attributes.}
In (g) and (h), when multiple entities of the same type with different attributes are present, both models also show degradation. This happens because, after embedding, the semantics of a black bird and a white bird may blend, thereby causing the erroneous generation. At the same time, we also notice that \textbf{our results are still better} than Structured Diffusion in these cases, as we still generate a black-and-white bird, whereas Structured Diffusion produced a bird with a giraffe's head, demonstrating an occurrence of entity leakage.

\section{Conclusion}

In this paper, we 
propose a training-free phase-wise attention control mechanism. We integrate novel temperature control within the self-attention module and phase-specific masking control in the cross-attention module. These attention controls enable model to more effectively shape image patches into coherent objects and significantly mitigate issues of entity fusion and misalignment. In our experiments, we evaluated our model using existing benchmark metrics and semi-human assessments tailored to distinct alignment scenarios. The experimental results demonstrate the robustness and efficacy of our model in alignment-focused image generation tasks.
\clearpage
\bibliographystyle{splncs04}
\bibliography{main}

\end{document}